%% file: main.tex
\documentclass[preprint,12pt]{elsarticle}

\usepackage{amssymb}

\usepackage{listings}
\usepackage{graphicx}
\usepackage{caption}
\usepackage{subcaption}

\newcommand{\joywilson}{Patient 1} 
\newcommand{\howard}{Patient 2} 
\newcommand{\steven}{Patient 3} 
\newcommand{\elizabeth}{Patient 4} 
\newcommand{\lizzy}{Patient 5}

\newcommand{\gdiez}{1 }
\newcommand{\gonce}{2 }
\newcommand{\gdoce}{3 }
\newcommand{\gtrece}{4 } 

\journal{Applied Soft Computing}

\begin{document}

\begin{frontmatter}



\title{Modeling Glycemia in Humans by Means of Grammatical Evolution}


\author[CES]{J. Manuel Colmenar}
\author[DACYA]{Jos\'{e} L. Risco-Martin}
\author[DACYA]{J. Ignacio Hidalgo}
\author[CES]{Alfredo Cuesta-Infante}
\author[HTOL]{Esther Maqueda}
\author[HUPA]{Marta Botella}
\author[HUPA]{Jos\'{e} Antonio Rubio}

\address[CES]{CES Felipe II, Univ. Complutense de Madrid (UCM) at Aranjuez, Spain}
\address[DACYA]{Dept. Arquitectura de Computadores y Autom\'{a}tica, UCM, Spain}
\address[HTOL]{Servicio de Endocrinolog\'{i}a y Nutrici\'{o}n (SEN), Hosp. Virgen de la Salud,Toledo, Spain}
\address[HUPA]{SEN, Hosp.Univ. Principe de Asturias, Alcal\'{a} de Henares, Spain}

\begin{abstract}
Diabetes mellitus is a disease that affects to hundreds of millions of people worldwide. Maintaining a good control of the disease is critical to avoid severe long-term complications. In recent years, several  artificial pancreas systems have been proposed and developed, which are increasingly advanced. However there is still a lot of research to do. One of the main problems that arises in the (semi) automatic control of diabetes, is to get a  model explaining how glycemia (glucose levels in blood) varies with insulin, food intakes and other factors, fitting  the characteristics of each individual or patient.
This paper proposes the application of evolutionary computation techniques  to obtain customized models of patients, unlike most of previous approaches which obtain averaged models. 
The proposal is based on a kind of genetic programming based on grammars known as Grammatical Evolution (GE). The proposal has been tested with in-silico patient data and results are clearly positive. We present also a study of four different grammars and five objective functions. In the test phase the models characterized the glucose with a mean percentage average error of 13.69\%, modeling  well also both hyper and hypoglycemic situations. 

\end{abstract}


\end{frontmatter}


\section{Introduction}
\label{sec:intro}
\input{intro.tex}

\section{Related Work}
\label{sec:related}
\input{relatedwork.tex}

\section{Evolutionary Approach}
\label{sec:gramevol}

\input{gramevol.tex}

\section{Model Description}
\label{sec:model}
\input{model.tex}

\section{Experimental Setup}
\label{sec:expsetup}
\input{expsetup.tex}

\section{Results}
\label{sec:results}
\input{results.tex}

\section{Conclusions and Future Work}
\label{sec:conclusions}
\input{conclusions.tex}





\bibliographystyle{elsarticle-num}
\bibliography{bioinspired}







\end{document}

%% file: intro.tex
Diabetes mellitus is a disease caused by a defect in either the secretion or in the action of insulin, which is essential for the control of blood glucose levels. 
Both of them cause in cells not to assimilate the sugar and, as a consequence, there is a rise in blood glucose levels, or hyperglycemia. Several types of diabetes differ in origin. According to the ADA (American Diabetes Association) we can distinguish four types of diabetes:

\begin{itemize}
\item Type 1 Diabetes (T1DM): Cells do not produce insulin because of an autoimmune process. Currently, requires the person to inject insulin or wear an insulin pump.
\item Type 2 Diabetes (T2DM): Results from insulin resistance, where cells fail to use insulin properly, sometimes combined with an absolute insulin deficiency.
\item Gestational Diabetes: appears in the gestation period in one out of ten pregnant women. Pregnancy is a change in the body's metabolism, since the fetus uses the mother's energy for food, oxygen and others. This causes a decrease in the secretion of insulin from the mother.
\item Other Types: such as problems on $\beta$ -cells, 
genetic defects affecting insulin action, induced by drugs, genetic syndroms, etc.
\end{itemize}

In most cases, diabetic patients with long time evolution need exogenous insulin either injected into various injection doses, or introduced by an insulin pump. It is important to maintain good glycemic control to prevent not only from the acute complications specific to diabetes (diabetic ketoacidosis and hypoglycemia, defined as blood glucose value less than $70 mg/dl$), but also from a set of multi-chronic complications associated with diabetic patients: nephropathy, retinopathy, microangiopathy and macroangiopathy. 



In recent years, it has been shown that a strict glycemic control in critically ill patients improves performance and reduces medical costs \cite{issc2008} \cite{Krinsley2006}.
Glucose levels control is a demanding and difficult task for both patients and their families. To keep good levels of blood glucose, the patient must 
have some capacity of prediction to know what level of glucose would have if ingested a certain amount of food or injected with a quantity of a insulin of a certain kind.
In fact, the objective is to avoid not only long periods of hyperglycemia (glucose levels $\geq 120 mg/dl$) but also episodes of severe hypoglycemia (glucose levels $\leq 40 mg/dl$) that can lead to patient death. 

One of the aspects that make it difficult to control blood glucose level is the lack of a general model of response to both insulin and the various factors mentioned above, due to the particularities of each patient \cite{Association2010}. Models in the literature apply classical modeling techniques, resulting in linear equations, defined profiles, or models with a limited set of inputs. 
Here we propose a novel technique that involves obtaining the patient model using genetic programming (GP). GP eliminates barriers in building the model, such as linearity or limitations on the input parameters.

Evolutionary techniques such as GP, have certain characteristics that make them particularly suited to address optimization problems and complex modeling. 
First, they are conceptually simple in its application but have a theoretical basis defined and widely studied. 
GP has demonstrated its applicability to many real problems, and is intrinsically parallelizable to work with a set of solutions. 
Furthermore, EAs have great potential to incorporate knowledge about the domain and to incorporate other search mechanisms (not necessarily evolutionary).

One of the best known applications of GP is symbolic regression and the application of one of its variants, Grammatical Evolution (GE), allows to obtain solutions that incorporate non-linear terms. GE is an evolutionary computation technique established in 1998 by Conor Ryan's group at the University of Limerick (Ireland) \cite{Ryan1998}. GP aims to find an executable program or function that respond to the reference data. The key advantage is that  GE applies genetic operators to a whole chain, which simplifies the search application in different programming languages. In addition, there are no memory problems, unlike with GP where the tree representation could have the well know problem of bloating (an excessive growing of the computer structures in memory). Hence, we propose to apply GE to find a custom model that describes and predicts the blood glucose level in a patient. Our method takes the historic data of a patient consisting in previous glucose levels, ingested carbohydrates and injected insulin, and obtains an expression that can be used to predict near future glucose values. 
The contributions of this work are: 
\begin{itemize}
\item We propose a method based on GE to obtain individualized and customized glycemia (glucose level in blood) models in humans.
\item We have tested this proposal with five in-silico patients taken from AIDA simulator \cite{AIDA}. 
 \item We present a study of four different grammars and five objective functions.
\item We have selected the best models for each patient and run a test phase with a new dataset. In the test phase the models characterized the glucose with a mean percentage average error of 13.69\%, reflecting also a good representation of both hyper and hypoglycemic situations. 
\end{itemize}
 
%
%

The rest of the paper is organized as follows. Section \ref{sec:related} describes the related work. Section \ref{sec:gramevol} details how grammatical evolution can be applied to this problem. Section \ref{sec:model} shows the general model we propose, as well as the grammars, particular models and objective functions we have studied for the glucose estimation problem. Section \ref{sec:expsetup} is devoted to the experimental setup, while Section \ref{sec:results} presents the results obtained in both training and test phases. Finally, Section \ref{sec:conclusions} explains the conclusions and the future work.

%% file: relatedwork.tex
Glucose level control is a very demanding and difficult task for both patients and their families. Trying to keep a good control of blood glucose involves tof perform blood glucose regular measurements (which involves at least one puncture in each measure or using a continuous monitoring system during some periods), insulin dose estimation, carbohydrates estimation, analyze that information somehow and to have some capacity of prediction that allows the patient to know what level of glucose would have if ingested a certain amount of food or injected with a quantity of a insulin of a certain kind.

As we have already mentioned, one of the main problems in controlling and predicting blood glucose levels is the lack of reliable models of response to both insulin and the various factors involved. Although there are some general approximations, there are hardly few adapted to the particularities of each patient \cite{Heusden2012a}\cite{Association2010}. The models in the literature apply classical modeling techniques, resulting in linear equations defined profiles, or models with a limited set of inputs. There are other factors that make a good control hard to achieve \cite{Cobelli2009}. For instance, we can mention that  there is a significant delay between insulin administration and the appearance of insulin in the blood stream with the use of subcutaneous (SC) insulin. This delay time  limits the achievable control performance on  subcutaneous administration of insulin.

In \cite{Heusden2012a} authors propose the use of models to maintain margins of robustness when there is a mismatch between the model and the patient. The approach used there is personalized using information a priori known (ie, easy access) of patients
to limit conservatism. However, this model only applies to linear models and can not incorporate other factors such as exercise or stress that clearly affect the 
expected levels of glucose. 
Models based on data for individual subjects are often inaccurate, since clinical data in T1DM are not extensive enough to identify the exact models \cite{Stahl2009}\cite{Finan2009}. To obtain continuous series of data, glucose levels should be measured using a  subcoutaneous continuous glucose monitoring (CGM) system.  To calculate the dose of insulin the patient or the physician may use different mechanisms and control algorithms. Hence, we can also find some personalized control approaches  \cite{Hovorka2010}\cite{El-Khatib2010}\cite{Magni2009}\cite{Kovatchev2010}
corresponding to clinical practice. Current treatment for subjects with T1DM uses rates of basal insulin delivery, insulin to carbohydrate ratios (CHO) and individual correction factors, typically from observations of the specialist.

There are also some models used in artificial pancreas systems or models of closed loop control  \cite{Dassau2009} \cite{Steil2011} . The main risk is hypoglycemia as a result of excessive insulin administration. However  we know that it is possible to reach a good control with approximate models, provided that the model is related to the control objective \cite{Gevers2005}\cite{Rivera1995}. Again, the most important factor for the focus of this paper is the lack of accurate individualized models. If there is an accurate model of the subject’s response to insulin, the design of the controller is relatively simple using classical control techniques. Autoregressive models (AR) may be applied to overcome problems of identifiability\cite{Gani2010}\cite{Sparacino2007}, although those are not useful for controlling since they have not an exogenous input. Some protocols have also been proposed to improve the reliability of the models \cite{Finan2009}\cite{Dassau2009} \cite{Lee2009} but the possibilities for the design of experiments are limited due to the strict security requirements and limitations in clinical protocols.
					
There have been also different approaches to facilitate the diabetes control from commercial companies. However, most of them have been designed only for specific glucometers and when providing insulin recommendations, the model is not available.  \emph{Glucofacts Deluxe} by \textit{Bayern}  \cite{BayernGLUCOFACTS}, \emph{CoPilot Health Management System} by  \textit{Abbot} \cite{Abbot2012}, and \emph{MenaDiab} \cite{Menari2012}  by  \textit{Menarini} are some of them.
	
Although there are many works that use control models, up to the date the modeling problem has not been addressed by evolutionary computation techniques that, as mentioned, have a high potential to incorporate to the model factors which are difficult to quantify, in other words to collect system dynamics. The main new aspect is the use of individualized models, i.e. we obtain a solution of the problem for each set of data on a single patient or individual. This approach has not been seen to date, given its complexity with traditional methods, but affordable with evolutionary methods.

%% file: gramevol.tex
The aim of this work is to find out an expression to model the glucose level of a diabetic patient. This expression should be obtained from  previous collected data of glucose, carbohydrates and insulin. Therefore, we deal with a kind of Symbolic Regression (SR) Problem. SR tries to obtain a mathematical expression to reproduce a set of discrete data.
Genetic Programming  GP has proven effective in a number of SR problems, although  there are some limitations, which often come in the way of representation.  such as bloating. Another point to be considered is that in GP, evolution is produced on the phenotype of the individual and not on its representation (genotype).
During last years, variants to GP like Grammatical Evolution (GE) appeared to propose different evaluation approaches.GE allows generation of computer programs in an arbitrary language. This is achieved by using grammars to specify the rules for obtaining the programs. Specifically we will use grammars expressed in Backus Naur form (BNF).





In contrast to genetic algorithms, which work with representation of solutions, GE works (evolves) with a genetic code that determines the production process of this solution. The code translation process is determined by  grammars represented as BNF.

BNF is a notation technique for expressing context-free grammars. The BNF can be any specification of a complete language or a subset of a problem-oriented language. A BNF specification is a set of derivation rules, expressed in the form:

 \[ \texttt{<symbol>\ ::=\ <expression>} \] 

The rules are composed of sequences of terminals and non-terminals. Symbols that appear at the left are non-terminals while terminals never appear on a left side. In this case we can affirm that \texttt{<symbol>}  is a non-terminal and, although this is not a complete BNF specification, we can affirm also that \texttt{<expression>} will be also a non-terminal since those are always enclosed between the pair \texttt{<>}. So, in this case the non-terminal \texttt{<symbol>} will be replaced (indicated by \texttt{::=})  by an expression. The rest of the grammar must indicate the different possibilities.

A grammar is represented by the 4-Tuple \{N, T, P, S\}, being N the non-terminal set, T is the terminal set, P the production rules for the assignment of elements on N and T, and  S is a start symbol which should appear in N. The options within a production rule are separated by a  "\texttt{|}" symbol.

Figure \ref{fig:SymbolicGrammar} represents an example of a grammar in BNF designed for symbolic regression. The code that represents an expression will consist of elements of the set of terminals T. These have been combined with the rules of the grammar, as will be  explained below.

\begin{figure}[!t]
\begin{lstlisting}[basicstyle=\footnotesize\ttfamily,breaklines=true,frame=tb]
N = { expr, op, pre_op ,var, num, dig }
T = { +, -, *, /, Sin, Cos, Abs, X, 0, 1, 2, 3, 4, 5, (, ),.}
S = { expr }
P = { I, II ,III ,IV ,V ,VI }

I    <expr>    ::=  <expr> <op> <expr>  
                  | <pre_op> (<expr>) 
                  | <var>

II   <op>      ::=  +  |  -  |  *  |  /
                 
III  <pre_op> ::=  Sin  |  Cos  |  Abs
  
IV   <var>     ::=  X  |  <num>
 
V    <num>    ::=  <dig>.<dig>  |  <dig>

VI   <dig>     ::=  0  |  1  |  2  |  3  |  4  |  5
\end{lstlisting}
\caption{Example of a grammar in BNF format designed for symbolic regression.}
\label{fig:SymbolicGrammar}
\end{figure}

Besides, grammars can be adapted to bias the search of the evolutionary process because there is a finite number of options on each production rule, which limits the search space.

\subsubsection*{Mapping Process}
\label{sec:MappingProcess}

As we have mentioned above we will use an EA to evolve genotypes, i.e. a string of integer values.  We use the individual genotype to map the start symbol onto terminals by reading codons which, in our work, have 8-bits length. The process is similar to the explained on the previous section, but instead of doing random choices, we will take our decisions by reading the individual genotype. Each codon  is represented by an integer value on the genotype, which is processed by the following mapping function:
\[Choice_i = (CIV)\  MOD\  (\#\ of\ choices_i)\]
where $Choice_i$ is the choice selected for non-terminal $i$, $CIV$ is the codon integer value we are decoding, $MOD$ is the module function, and $(\#\ of\ choices_i)$ is the number of possible choices at rule for the non-terminal $i$.

The mapping function was proposed in \cite{Ryan1998} and takes the integer value of the chromosome, computes the module function in relation to the number of the choices of a rule, and selects the choice according to that result. Given that the module function will return values from $0$ to $(\#\ of\ choices_i)-1$, the first choice will correspond to the first value, $0$, the second to $1$, and so on. Therefore, if a rule has only one possible choice, this choice will always be selected because $k\  MOD\  1 = 0$ for any $k$ integer value.


We will illustrate the mapping process using the example grammar shown in Figure \ref{fig:SymbolicGrammar}, designed for solving a symbolic regression problem, which is indeed a possible grammar for the glucose model problems (we only need to particularize the terminal set as it will be explained on next section). An individual is composed of a set of integer genes. Each gene can take a numeric value from 0 to 255 since we are working with codons of 8 bits. Let us suppose we are mapping the following 7-genes individual:
\[ 12-55-23-47-38-254-2 \]

The start symbol is \texttt{S =  \{ expr \}}, hence the solution expression will begin with this non-terminal:

\[Solution =\ \texttt{<expr>} \]

Now, in order to obtain the phenotype, we will apply the mapping function on the first gene ($CIV=12$) using the rule for first non-terminal of the expression. At this point only one non-terminal appears, \texttt{<expr>}, which corresponds to rule \texttt{I} of Figure \ref{fig:SymbolicGrammar}. The number of choices in that rule is 3. Hence, the mapping is applied:
\[12\ MOD\ 3 = 0\]
so, we select the first option, \texttt{<expr> <op> <expr>}, and continue with the mapping. The selected option substitutes the decoded non-terminal. As a consequence, the current expression is the following:

\[Solution =\ \texttt{<expr> <op> <expr>} \]

The process will continue with the next codon, 55, which is used to decode the first non-terminal of the current expression, namely, \texttt{<expr>}. Again, we apply the mapping function on rule \texttt{I}:
\[55\ MOD\ 3 = 1\]
the second option, \texttt{<pre\_op> (<expr>)}, is selected, and the current expression is

\[Solution = \texttt{<pre\_op> (<expr>) <op> <expr>} \]

The next gene, 23, is now taken for decoding. Notice that in this point of the process, the first non-terminal that appears in the expression is \texttt{<pre\_op>}. Therefore, we apply the mapping function on rule \texttt{III}, which also has 3 possible choices:
\[23\ MOD\ 3 = 2\]
Value 2 means to select the third option, the terminal symbol $Abs$. The resulting expression is

\[Solution = Abs(\texttt{<expr>})\ \texttt{<op> <expr>} \]

Next codon, 47, decodes \texttt{<expr>} with rule \texttt{I}:
\[47\ MOD\ 3 = 2\]
Value 2 means to select the third option, \texttt{<var>}. The resulting expression is

\[Solution = Abs(\texttt{<var>})\ \texttt{<op> <expr>} \]

Gene 38 decodes  \texttt{<var>} with rule \texttt{IV}:
\[38\ MOD\ 2 = 0\]
This value selects the first option, non-terminal \texttt{X}.

\[Solution = Abs(X)\ \texttt{<op> <expr>} \]

Non-terminal \texttt{<op>} is decoded with 254 and rule \texttt{II}:
\[254\ MOD\ 4 = 2\]
This value selects the third option, terminal \texttt{*}.

\[Solution = Abs(X)\ *\ \texttt{<expr>} \]

Next codon, 2, decodes \texttt{<expr>} with rule \texttt{I}:
\[2\ MOD\ 3 = 2\]
This value selects the third option, non-terminal \texttt{<var>}.

\[Solution = Abs(X)\ *\ \texttt{<var>} \]

At this point, the genotype-to-phenotype process has run out of codons. That is, once we have used all the genes or codons we have not arrived to an expression with terminals in all of its components. 

The solution is to reuse codons starting from the first one, although this is not usual in other EA approaches. In fact it is possible to reuse the codons more than once. This technique is known as \textit{wrapping} and mimics the gene-overlapping phenomenon of many organisms \cite{ONeill2003}. Reusing codons it is not a problem since in GE a codon always generates the same integer value and, if applied to the same rule, it generates the same solutions. However, if we use it with different rules we will obtain different phenotypes parts. What the GE grammars should make certain is that an individual genotype will always produce the same phenotype. In these conditions wrapping is not a problem.

So, applying wrapping, the process go back to first gene, 12, which is used to decode \texttt{<var>} with rule \texttt{IV}:
\[12\ MOD\ 2 = 0\]
This value selects the first option, non-terminal $X$, giving the final expression of the phenotype.

\[Solution = Abs(X)\ *\ X \]


In the next section we describe how the four grammars under study represent different search spaces for expressions to model the blood glucose level.

%% file: model.tex
A model for glucose levels should be based on observable factors as well as on intrinsic non-observable features of the patient's body. Observable factors are those data that either the patient or a measure machine can collect, while non-observable factors should be inferred. Hence, we propose a model that considers all these factors, applying GE to infer an expression that characterize the behavior of the glucose in diabetic patients. In addition, we describe in this section the different objective functions we have studied to make GE evolve towards useful expressions for the model.

\subsection{Available Data and General Glucose Model}

The actual level of glucose in the patient's blood depends on several factors, some of them intrinsic to its own organism functions \cite{Man2006}. The most important among these factors are the glucose level, the carbohydrates ingested and the insulin injected.

These factors are considered in the datasets of our in-silico patients, which were obtained with AIDA simulator \cite{AIDA}. Notice that, for real patients, these data are easy to collect. Actual glucose values are obtained from blood analyzers, carbohydrate units ingested are calculated based on the daily meals, and insulin injected, distinguished by insulin type, is also an information that the patient usually knows.

Therefore, we have developed our research based on collections of data that follow the previous idea. More precisely, our data series represent measures taken each 15 minutes along the day. Table \ref{tab:datosmodelo} shows a 24-hours dataset of one of our in-silico patients, named \joywilson. For each time step, represented in one line of the table, $k$ is the actual time, $GL$ is the actual glucose level, $CH$ is the carbohydrates units ingested, $IS$ is the short effect insulin injected and $IL$ is the long effect insulin injected.

\begin{table}\tiny
\begin{tabular}{|c|c|c|c|c|}
\hline
k & GL & CH & IS & IL \\
\hline
1 & 209.1453 & 0 & 0 & 0 \\
2 & 209.1453 & 0 & 0 & 0 \\
3 & 205.79354 & 0 & 0 & 0 \\
4 & 202.56395 & 0 & 0 & 0 \\
5 & 199.43946 & 0 & 0 & 0 \\
6 & 196.41567 & 0 & 0 & 0 \\
7 & 193.49677 & 0 & 0 & 0 \\
8 & 190.69246 & 0 & 0 & 0 \\
9 & 188.01547 & 0 & 0 & 0 \\
10 & 185.47971 & 0 & 0 & 0 \\
11 & 183.09885 & 0 & 0 & 0 \\
12 & 180.88533 & 0 & 0 & 0 \\
13 & 178.84963 & 0 & 0 & 0 \\
14 & 176.9999 & 0 & 0 & 0 \\
15 & 175.34167 & 0 & 0 & 0 \\
16 & 173.87786 & 0 & 0 & 0 \\
17 & 172.60884 & 0 & 0 & 0 \\
18 & 171.53263 & 0 & 0 & 0 \\
19 & 170.6451 & 0 & 0 & 0 \\
20 & 169.94031 & 0 & 0 & 0 \\
21 & 169.41076 & 0 & 0 & 0 \\
22 & 169.04771 & 0 & 0 & 0 \\
23 & 168.84148 & 0 & 0 & 0 \\
24 & 168.78173 & 0 & 0 & 0 \\
25 & 168.85766 & 0 & 0 & 0 \\
26 & 169.05827 & 0 & 0 & 0 \\
27 & 169.37254 & 0 & 0 & 0 \\
28 & 169.78957 & 0 & 0 & 0 \\
29 & 170.29872 & 0 & 0 & 0 \\
30 & 170.88974 & 0 & 0 & 0 \\
31 & 171.55425 & 0 & 0 & 0 \\
32 & 172.27976 & 0 & 3 & 12 \\
33 & 173.05923 & 30 & 0 & 0 \\
\hline 
\end{tabular}
\begin{tabular}{|c|c|c|c|c|}
\hline
k & GL & CH & IS & IL \\
\hline
34 & 174.09018 & 0 & 0 & 0 \\
35 & 176.68898 & 0 & 0 & 0 \\
36 & 184.17896 & 0 & 0 & 0 \\
37 & 195.61325 & 0 & 0 & 0 \\
38 & 209.11478 & 0 & 0 & 0 \\
39 & 223.51534 & 0 & 0 & 0 \\
40 & 237.54628 & 0 & 0 & 0 \\
41 & 247.25104 & 20 & 0 & 0 \\
42 & 250.72465 & 0 & 0 & 0 \\
43 & 251.90543 & 0 & 0 & 0 \\
44 & 255.86031 & 0 & 0 & 0 \\
45 & 262.91532 & 0 & 0 & 0 \\
46 & 271.63911 & 0 & 0 & 0 \\
47 & 277.89942 & 0 & 0 & 0 \\
48 & 278.70943 & 0 & 0 & 0 \\
49 & 275.38173 & 40 & 0 & 0 \\
50 & 269.9759 & 0 & 0 & 0 \\
51 & 264.80613 & 0 & 0 & 0 \\
52 & 264.5835 & 0 & 0 & 0 \\
53 & 269.03811 & 0 & 0 & 0 \\
54 & 276.56921 & 0 & 0 & 0 \\
55 & 286.01609 & 0 & 0 & 0 \\
56 & 296.54932 & 0 & 0 & 0 \\
57 & 307.58483 & 0 & 0 & 0 \\
58 & 317.78442 & 0 & 0 & 0 \\
59 & 322.9634 & 0 & 0 & 0 \\
60 & 322.05895 & 0 & 0 & 0 \\
61 & 317.01482 & 0 & 0 & 0 \\
62 & 309.34854 & 0 & 0 & 0 \\
63 & 300.16725 & 0 & 0 & 0 \\
64 & 290.26909 & 10 & 0 & 0 \\
65 & 280.21928 & 0 & 0 & 0 \\
\hline 
\end{tabular}
\begin{tabular}{|c|c|c|c|c|}
\hline
k & GL & CH & IS & IL \\
\hline
66 & 272.82541 & 0 & 0 & 0 \\
67 & 271.4209 & 0 & 0 & 0 \\
68 & 272.60383 & 0 & 0 & 0 \\
69 & 271.58261 & 0 & 0 & 0 \\
70 & 268.30335 & 0 & 0 & 0 \\
71 & 263.63006 & 0 & 4 & 0 \\
72 & 258.18738 & 0 & 0 & 0 \\
73 & 252.2917 & 30 & 0 & 0 \\
74 & 246.13977 & 0 & 0 & 0 \\
75 & 240.98451 & 0 & 0 & 0 \\
76 & 241.19843 & 0 & 0 & 0 \\
77 & 246.19998 & 0 & 0 & 0 \\
78 & 254.31449 & 0 & 0 & 0 \\
79 & 264.42406 & 0 & 0 & 0 \\
80 & 275.21267 & 0 & 0 & 0 \\
81 & 282.61864 & 0 & 0 & 0 \\
82 & 284.42698 & 0 & 0 & 0 \\
83 & 282.33176 & 0 & 0 & 0 \\
84 & 277.80323 & 0 & 0 & 0 \\
85 & 271.88307 & 0 & 0 & 0 \\
86 & 265.29768 & 0 & 0 & 0 \\
87 & 258.54258 & 0 & 0 & 0 \\
88 & 251.94544 & 0 & 0 & 0 \\
89 & 245.71299 & 0 & 0 & 0 \\
90 & 239.91877 & 0 & 0 & 0 \\
91 & 234.68683 & 0 & 0 & 18 \\
92 & 230.03136 & 0 & 0 & 0 \\
93 & 225.93797 & 0 & 0 & 0 \\
94 & 222.23379 & 0 & 0 & 0 \\
95 & 218.78458 & 0 & 0 & 0 \\
96 & 215.4919 & 0 & 0 & 0 \\
97 & 212.29133 & 0 & 0 & 0 \\
\hline 
\end{tabular}
	\caption{24-hours dataset for in-silico patient \joywilson.}
	\label{tab:datosmodelo}
\end{table}

In our dataset $k$ represents the time step corresponding to a moment of the day. Then, $k=1$ represents 12:00 AM, $k=2$ represents 12:15 AM, and so on. As seen in the table, many of the time steps do not have any data about carbohydrates or insulin, whereas time steps surrounding the meal hours do provide that information.

The model we propose provides estimated glucose values, denoted as $\widehat{GL}$. Hence, for each time step, estimated glucose is obtained by using previous estimated glucose values and actual carbohydrates and insulin units.  A general form of this model should be similar to the following:

\begin{equation}
\label{eq:modelogenerico}
\widehat{GL}(k+1) = f(\widehat{GL},CH,IS,IL), 1 \le k \le N
\end{equation}

where $\widehat{GL}(k+1)$ is the next estimated glucose value, $\widehat{GL}$ corresponds to previous  estimated glucose values, $CH$ corresponds to previously ingested carbohydrates and $IS$ and $IL$ correspond to previously injected insulin for both types, short and long effect. Therefore, the dataset provides input values for the variables in our glucose model proposal.

In this way, the GE engine should be able to decide how $f$ looks like. However, in order to guide the search of the evolutionary process, we do need a grammar that will both limit the search space and represent the behavior of the blood glucose level. Next, we detail the grammars that we studied in this work.

\subsection{BNF Grammars for Modelling Glucose Levels}
\label{sec:BNFGrammarsForModellingGlucoseLevels}

Following the general model shown in (\ref{eq:modelogenerico}), we have designed four grammars where the estimated glucose depends on the observable factors. As shown in \cite{Hemberg2013} the incorporation of some of the problem's knowledge into the grammar will improve the exploration performance. Therefore, we designed an expression for glucose which depends on previous glucose, carbohydrates and insulin. This expression is coded as rule \texttt{I} in all our grammars but, as seen next, is surrounded by different rules that are translated into different concrete models.

The grammars were designed by following the advice of the medical doctors in our research team. According to them, the expected behavior of the glucose depends on previous carbohydrates ingested and insulin injected, but it may vary along the day in a different way for each patient. In addition, glucose may be influenced at different degrees by each ingestion of carbohydrates and each injection of insulin. Therefore, we selected four different approaches that considered different degrees of influence, as well as different influence time windows.

\subsubsection*{Grammar \gdiez}
Given that it is well known that carbohydrate ingestion rises glucose while insulin injections lowers it, we tried a grammar with such a behavior. The general model will be approximated with expressions similar to (\ref{eq:modelog10}), where any previous values of glucose, carbohydrates and insulin may be used. Besides, carbohydrates are always added, while insulin values are always subtracted. 
\begin{equation}\footnotesize
\label{eq:modelog10}
\widehat{GL}(k+1) = f_{gl}(\widehat{GL}(k-m)) + f_{ch}(CH(k-m)) - f_{in}(IS(k-m),IL(k-m)), 0 \le m \le k
\end{equation}

The concrete form of $f_{gl}$, $f_{ch}$ and $f_{in}$ will be determined by GE with the help of the grammar that we called Grammar \gdiez, shown in Figure \ref{fig:Grammar10}. The three terms \texttt{<exprgluc>}, \texttt{<exprch>} and \texttt{<exprins>} correspond to $f_{gl}$, $f_{ch}$ and $f_{in}$, respectively, and they are expressions that could use prefix operands like those in rule \texttt{IX}, variables for each of one the terms, or combinations of them through operators in rule \texttt{VIII}.

\begin{figure}[ht]
\begin{lstlisting}[basicstyle=\scriptsize\ttfamily,breaklines=true,frame=tb]
N = {func, exprgluc, gluc, exprch, varch, exprins, varins, op, preop, idx, cte, dgt}
T = { +,-,*,/, sin, cos, tan, exp, 0, 1, 2, 3, 4, 5, 6, 7, 8, 9, 0, GL, CH, IS, IL, K}
S = {func}
P = {I, II ,III ,IV ,V ,VI ,VII, VIII, IX, X, XI, XII}

I    <func> ::= <exprgluc> + <exprch> - <exprins>

II   <exprgluc> ::= <preop> (<gluc>)
	|(<cte> <op> <gluc>)
	|<gluc>
	
III  <gluc> ::= #{GL[k_<idx>]}|#{K}

IV   <exprch> ::= <exprch> <op> <exprch>
	|<preop> (<exprch>)
	|<varch>

V    <varch> ::= #{CH[k_<idx>]}|#{K}|<cte>

VI   <exprins> ::= <exprins> <op> <exprins>
	|<preop> (<exprins>)
	|<varins>

VII  <varins> ::= #{IS[k_<idx>]}|#{IL[k_<idx>]}|#{K}|<cte>

VIII <op> ::=+|-|/|*
IX   <preop>::=sin|cos|tan|exp
X    <idx> ::= <dgt><dgt>
XI   <cte> ::= <dgt><dgt>.<dgt><dgt>
XII  <dgt>::=0|1|2|3|4|5|6|7|8|9
\end{lstlisting}
\caption{Grammar \gdiez. Any previous carbohydrates and insulin; carbohydrates are added and insulin subtracted.}
\label{fig:Grammar10}
\end{figure}

\subsubsection*{Grammar \gonce}
This grammar is a particularization of the previous one in the sense that it does not allow any previous value of variables. On the contrary, the grammar limits the values to just the two previous data in time, that is, $k$ and $k-1$. The resulting model, with the only difference of the range allowed for $m$, is shown in (\ref{eq:modelog11}).

\begin{equation}\footnotesize
\label{eq:modelog11}
\widehat{GL}(k+1) = f_{gl}(\widehat{GL}(k-m)) + f_{ch}(CH(k-m)) - f_{in}(IS(k-m),IL(k-m)), 0 \le m \le 1
\end{equation}

Figure \ref{fig:Grammar11} shows the grammar, where the indexes are limited to $00$ and $01$ in rules \texttt{III}, \texttt{V} and \texttt{VII}, which means the current and previous values of each variable.

\begin{figure}[ht]
\begin{lstlisting}[basicstyle=\scriptsize\ttfamily,breaklines=true,frame=tb]
N = {func, exprgluc, gluc, exprch, varch, exprins, varins, op, preop, cte, dgt}
T = { +,-,*,/, sin, cos, tan, exp, 0, 1, 2, 3, 4, 5, 6, 7, 8, 9, 0, GL, CH, IS, IL, K}
S = {func}
P = {I, II ,III ,IV ,V ,VI ,VII, VIII, IX, X, XI}

I    <func> ::= <exprgluc> + <exprch> - <exprins>

II   <exprgluc> ::= <preop> (<gluc>)
	|(<cte> <op> <gluc>)
	|<gluc>
	
III  <gluc> ::= #{GL[k_00]}|#{GL[k_01]}|#{K}

IV   <exprch> ::= <exprch> <op> <exprch>
	|<preop> (<exprch>)
	|<varch>

V    <varch> ::= #{CH[k_00]}|#{CH[k_01]}|#{K}|<cte>

VI   <exprins> ::= <exprins> <op> <exprins>
	|<preop> (<exprins>)
	|<varins>

VII  <varins> ::= #{IS[k_00]}|#{IS[k_01]}|#{IL[k_00]}|#{IL[k_01]}|#{K}|<cte>

VIII <op> ::=+|-|/|*
IX   <preop>::=sin|cos|tan|exp
X    <cte> ::= <dgt><dgt>.<dgt><dgt>
XI   <dgt>::=0|1|2|3|4|5|6|7|8|9
\end{lstlisting}
\caption{Grammar \gonce. Only two previous values for carbohydrates and insulin are allowed; carbohydrates are added and insulin subtracted.}
\label{fig:Grammar11}
\end{figure}

\subsubsection*{Grammar \gdoce}
In order to provide more freedom to the search, we decided to leave the connecting operands opened to any simple arithmetic operation. Therefore, the model changes as shown in (\ref{eq:modelg12}), and $f$ corresponds to the function that connects the three expressions.

\begin{equation}\scriptsize
\label{eq:modelg12}
\widehat{GL}(k+1) = f(f_{gl}(\widehat{GL}(k-m)), f_{ch}(CH(k-m)), f_{in}(IS(k-m),IL(k-m))), 0 \le m \le k
\end{equation}

The grammar that defines this model is Grammar \gdoce, which presents a slight modification of the rule \texttt{I} of Grammar \gdiez. It consists on changing the fixed $+$ and $-$ operands with the non-terminal \texttt{<op>}, which can be any of the four arithmetic operands in rule \texttt{VIII}. Figure \ref{fig:Grammar12} shows the grammar.

\begin{figure}[ht]
\begin{lstlisting}[basicstyle=\scriptsize\ttfamily,breaklines=true,frame=tb]
N = {func, exprgluc, gluc, exprch, varch, exprins, varins, op, preop, idx, cte, dgt}
T = { +,-,*,/, sin, cos, tan, exp, 0, 1, 2, 3, 4, 5, 6, 7, 8, 9, 0, GL, CH, IS, IL, K}
S = {func}
P = {I, II ,III ,IV ,V ,VI ,VII, VIII, IX, X, XI, XII}

I    <func> ::= <exprgluc> <op> <exprch> <op> <exprins>

II   <exprgluc> ::= <preop> (<gluc>)
	|(<cte> <op> <gluc>)
	|<gluc>
	
III  <gluc> ::= #{GL[k_<idx>]}|#{K}

IV   <exprch> ::= <exprch> <op> <exprch>
	|<preop> (<exprch>)
	|<varch>

V    <varch> ::= #{CH[k_<idx>]}|#{K}|<cte>

VI   <exprins> ::= <exprins> <op> <exprins>
	|<preop> (<exprins>)
	|<varins>

VII  <varins> ::= #{IS[k_<idx>]}|#{IL[k_<idx>]}|#{K}|<cte>

VIII <op> ::=+|-|/|*
IX   <preop>::=sin|cos|tan|exp
X    <idx> ::= <dgt><dgt>
XI   <cte> ::= <dgt><dgt>.<dgt><dgt>
XII  <dgt>::=0|1|2|3|4|5|6|7|8|9
\end{lstlisting}
\caption{Grammar \gdoce. Any previous carbohydrates and insulin; connector operators selected from rule \texttt{VIII}.}
\label{fig:Grammar12}
\end{figure}

\subsubsection*{Grammar \gtrece}
The model here is the same as in Grammar \gonce, but giving freedom to operands that connect the expressions for glucose, carbohydrates and insulin, as done in Grammar \gdoce. The model is shown in (\ref{eq:modelg13}), where $f$ corresponds to the function that connects the three expressions.

\begin{equation}\scriptsize
\label{eq:modelg13}
\widehat{GL}(k+1) = f(f_{gl}(\widehat{GL}(k-m)), f_{ch}(CH(k-m)), f_{in}(IS(k-m),IL(k-m))), 0 \le m \le 1
\end{equation}

Therefore, Grammar \gtrece$ $  is similar to Grammar \gonce, but giving freedom to operands in rule \texttt{I}. Figure \ref{fig:Grammar13} shows the grammar.

\begin{figure}[ht]
\begin{lstlisting}[basicstyle=\scriptsize\ttfamily,breaklines=true,frame=tb]
N = {func, exprgluc, gluc, exprch, varch, exprins, varins, op, preop, cte, dgt}
T = { +,-,*,/, sin, cos, tan, exp, 0, 1, 2, 3, 4, 5, 6, 7, 8, 9, 0, GL, CH, IS, IL, K}
S = {func}
P = {I, II ,III ,IV ,V ,VI ,VII, VIII, IX, X, XI}

I    <func> ::= <exprgluc> <op> <exprch> <op> <exprins>

II   <exprgluc> ::= <preop> (<gluc>)
	|(<cte> <op> <gluc>)
	|<gluc>
	
III  <gluc> ::= #{GL[k_00]}|#{GL[k_01]}|#{K}

IV   <exprch> ::= <exprch> <op> <exprch>
	|<preop> (<exprch>)
	|<varch>

V    <varch> ::= #{CH[k_00]}|#{CH[k_01]}|#{K}|<cte>

VI   <exprins> ::= <exprins> <op> <exprins>
	|<preop> (<exprins>)
	|<varins>

VII  <varins> ::= #{IS[k_00]}|#{IS[k_01]}|#{IL[k_00]}|#{IL[k_01]}|#{K}|<cte>

VIII <op> ::=+|-|/|*
IX   <preop>::=sin|cos|tan|exp
X    <cte> ::= <dgt><dgt>.<dgt><dgt>
XI   <dgt>::=0|1|2|3|4|5|6|7|8|9
\end{lstlisting}
\caption{Grammar \gtrece. Only two previous values for carbohydrates and insulin are allowed; connector operators selected from rule \texttt{VIII}.}
\label{fig:Grammar13}
\end{figure}

Once the grammars are presented, we next describe the fitness evaluation of an individual, as well as the different objectives studied in this work.

\subsection{Fitness Evaluation}
Grammars limit the search space in GE, but fitness functions are committed to guide the evolution to a good solution. So, in order to obtain the fitness of an individual, our evolutionary process first obtains the glucose values generated by the expression of the individual phenotype. As explained before, these are the estimated glucose values, denoted as $\widehat{GL}$. So, for each time step, estimated glucose is obtained by using previous estimated glucose values and actual carbohydrates and insulin units.

Once $\widehat{GL}$ is obtained, the absolute difference between the actual and the predicted glucose values is calculated for each time step. As in the general symbolic regression problem, this measure is called the error. The formula we apply is shown in (\ref{eq:error}), where $GL$ is the actual glucose value and $\widehat{GL}$ is the glucose value that the phenotype expression generates. We have studied five different objectives with their corresponding fitness functions, shown in Table \ref{tab:functions}.

\begin{equation}
\label{eq:error}
e_k=|GL(k)-\widehat{GL}(k)|, 1 \le k \le N
\end{equation}


\begin{table}
\begin{center}
\begin{tabular}{| c | c |}
\hline
Objective & Fitness Function \\
\hline

Least Squares & $F_1=\sum^{N}_{k=1} {{e_k}^2}$ \\[2ex]

Average Error & $F_2=\frac{1}{N}\sum^{N}_{k=1} {e_k}$ \\[2ex]

Maximum Error & $F_3=max({e_k}), 1 \le k \le N$ \\[1.5ex]

RSME & $F_4=\sqrt{\frac{1}{N}\sum^{N}_{k=1} {{e_k}^2}}$ \\[2ex]

MAD & $F_5=\frac{1}{N}\sum^{N}_{k=1} \frac{e_k}{GL(k)}$ \\[2ex]

\hline 
\end{tabular}
\end{center}
	\caption{Fitness functions for the five objectives under study. $N$ is the total number of measures.}
	\label{tab:functions}
\end{table}

%% file: expsetup.tex
In this section we describe the characteristics of the five in-silico patients we dealt with, as well as the configuration of each set of experiments.

\subsection{In-silico Patients}
We work with a set of in-silico patients obtained with AIDA simulator \cite{AIDA}. The website of the simulator offers several characterized patients from which we selected five of them. The glucose values for each patient were obtained by introducing different carbohydrates and insulin values and then running the simulator. The description of each one of the patients can be found on the website, but we replicate them here for the sake of clarity. The patients are the following:

\textbf{\joywilson}. This woman is on three injections of short and/or intermediate acting insulin each day, with a split-evening dose.  She wants to start a family, but consistently has had quite high blood glucose levels in the early afternoon.

\textbf{\howard}. This 45 year old man was diagnosed as having diabetes at the age of 14. He is currently on a regimen of combined short and/or intermediate acting insulin preparations four times per day.  As you can see from his home monitoring blood glucose measurements, he tends to higher blood glucose values overnight but has a low blood glucose in the mid-morning.

\textbf{\steven}. This man is a relatively newly diagnosed insulin-dependent (type 1) diabetic patient.  He has had problems maintaining his blood glucose profile on two and more recently three injections per day; so currently he is controlled on four injections per day.  He tends to quite high blood glucose levels in the middle of the day, despite not eating excessively. 

\textbf{\elizabeth}. It has taken a lot of effort to stabilize this girl's blood glucose profile. However, she still often goes hypoglycemic in the middle of the day, especially between breakfast and lunch.  She is on a slightly unusual regimen taking a short acting insulin preparation three times per day, with an intermediate acting preparation twice a day -- at lunchtime and before bed.

\textbf{\lizzy}. This overweight 58 year old insulin-dependent (type 1) diabetic patient has had major problems losing weight.  She is quite sensitive to insulin. In addition, she smokes and is at great risk of suffering a heart attack or stroke.

\subsection{Genetic Parameters}

As with genetic programming, GE can use any search algorithm able to operate on integer or binary strings. We have selected a simple GA with one-point crossover and point mutation. Population initialization is made by randomly generating fixed integer strings. Table \ref{tab:geneticparams} shows the rest of the genetic and GE parameters.

\begin{table}
\begin{center}
\begin{tabular}{|l|c|}
\hline
Parameter & Value \\
\hline
Max. wraps & 3 \\
Codon size & 256 \\		
Chromosome length & 100 \\
Population size & 100 \\
Generations & 2500 \\
Crossover probability & 0.6 \\
Mutation probability & 0.2 \\
Tournament size & 2 \\
\hline 
\end{tabular}
\end{center}
	\caption{Parameters for GE experiments.}
	\label{tab:geneticparams}
\end{table}

%% file: results.tex
Our experiments are divided into training and test phases. The objective of the training phase is to evaluate the performance of the proposed grammars in combination with the fitness functions, as well as to obtain models that characterize the glucose behavior on each patient. In this phase, the training dataset is formed by the 24-hours records of five in-silico patients. We have executed 30 runs with the same configuration of grammar and objective for each patient. Given that we have studied four grammars and five different objectives, a total of 600 runs were performed for each one of the in-silico patients in the training phase. Hence, we have obtained 600 glucose models for each patient.

In order to validate the goodness of the models, we have performed the test phase, where no GE was applied. In this phase, a different set of 24-hours records was employed for the same five in-silico patients. Using this test dataset, we have calculated the glucose values of each patient applying the best models obtained in the training phase.




Next, we analyze the optimizations and describe the results obtained in both phases.

\subsection{Training Phase}
In order to compare the performance of the grammars, we have obtained the average fitness for each set of optimization runs. We have grouped the runs by objective function, comparing the results for all patients. 

Table \ref{tab:ResultsObj0} shows the mean and standard deviation fitness values for least squares objective. As seen, Grammar \gonce obtains the best average fitness values for three of the patients, and is very close to the best value for \lizzy. Grammar \gtrece obtains values close to Grammar \gonce, but always worse.

\begin{table}\scriptsize
\begin{center}
\begin{tabular}{|l|c|c|c|c|}
\hline
Patient & Grammar \gdiez & Grammar \gonce & Grammar \gdoce & Grammar \gtrece \\ 
\hline 
\joywilson & $90707.25_{34144.06}$ & $\mathbf{45248.19_{8007.10}}$ & $107420.94_{22109.20}$ & $45595.31_{7059.14}$ \\ 
\howard & $178148.92_{8511.68} $ & $\mathbf{163723.13_{92492.90}} $ & $189369.64_{49435.68} $ & $172541.57_{94798.73}$ \\  
\steven & $83291.12_{20894.22} $ & $\mathbf{50788.42_{10489.82}} $ & $95872.31_{24151,13} $ & $60035.69_{16633.91}$ \\
\elizabeth & $89494.99_{10067.22} $ & $97425.94_{11745.69} $ & $\mathbf{87620.73_{11431.03}} $ & $98039.99_{12244.67}$ \\
\lizzy & $\mathbf{46531.15_{10810.73}} $ & $46826.41_{9258.46} $ & $49618.53_{10649.51} $ & $49502.43_{13359.12}$ \\
\hline 
\end{tabular}
	\caption{Mean and standard deviation fitness values of $F_1$, \textbf{least squares}.}
	\label{tab:ResultsObj0}

\begin{footnotesize}
\vspace{2em}
\begin{tabular}{|l|c|c|c|c|}
\hline
Patient & Grammar \gdiez & Grammar \gonce & Grammar \gdoce & Grammar \gtrece \\ 
\hline 
\joywilson & $25.74_{4.38} $ & $16.82_{1.82} $ & $25.72_{3.95} $ & $\mathbf{16.64_{1.73}}$ \\ 
\howard & $30.96_{1.46} $ & $34.35_{8.27} $ & $31.06_{2.25} $ & $\mathbf{29.66_{9.04}}$ \\  
\steven & $24.86_{3.30} $ & $ \mathbf{18.09_{2.47}} $ & $24.85_{4.25} $ & $19.19_{3.36}$ \\
\elizabeth & $23.94_{1.82} $ & $25.18_{2.19} $ & $\mathbf{23.50_{1.52}} $ & $24.54_{2.37}$ \\
\lizzy & $\mathbf{16.39_{2.02}} $ & $17.13_{1.76} $ & $16.74_{1.92} $ & $17.28_{1.75}$ \\
\hline 
\end{tabular}
	\caption{Mean and standard deviation fitness values of $F_2$, \textbf{average error}.}
	\label{tab:ResultsObj1}

\vspace{2em}
\begin{tabular}{|l|c|c|c|c|}
\hline
Patient & Grammar \gdiez & Grammar \gonce & Grammar \gdoce & Grammar \gtrece \\ 
\hline 
\joywilson & $68.29_{7.99} $ & $43.74_{1.72} $ & $67.99_{9.72} $ & $\mathbf{42.94_{1.87}}$ \\ 
\howard & $105.24_{9.56} $ & $98,87_{18.35} $ & $\mathbf{98.09_{11.95}} $ & $106.20_{14.38}$ \\  
\steven & $54.95_{3.47} $ & $\mathbf{44.38_{3.70}} $ & $56.82_{4.67} $ & $44.40_{3.37}$ \\
\elizabeth & $67.92_{3.37} $ & $68.52_{3.04} $ & $\mathbf{66.14_{5.32}} $ & $68.49_{3.14}$ \\
\lizzy & $48.23_{3.71} $ & $44.07_{3.60} $ & $46.83_{5.10} $ & $\mathbf{43.80_{6.55}}$ \\
\hline 
\end{tabular}
	\caption{Mean and standard deviation fitness values of $F_3$, \textbf{maximum error}.}
	\label{tab:ResultsObj2}

\vspace{2em}
\begin{tabular}{|l|c|c|c|c|}
\hline
Patient & Grammar \gdiez & Grammar \gonce & Grammar \gdoce & Grammar \gtrece \\ 
\hline 
\joywilson & $31.52_{5.22} $ & $\mathbf{21.20_{1.35}} $ & $33.04_{3.77} $ & $21.67_{2.08}$ \\ 
\howard & $42.51_{1.57} $ & $\mathbf{41.85_{11.17}} $ & $42.20_{4.90} $ & $44.61_{10.60}$ \\  
\steven & $29.29_{3.91} $ & $\mathbf{22.50_{2.81}} $ & $30.03_{4.28} $ & $23.64_{3.35}$ \\
\elizabeth & $30.71_{1.68} $ & $31.22_{2.44} $ & $\mathbf{29.56_{2.05}} $ & $30.37_{2.82}$ \\
\lizzy & $22.45_{1.96} $ & $\mathbf{21.54_{2.40}} $ & $22.48_{2.74} $ & $21.59_{3.02}$ \\
\hline 
\end{tabular}
	\caption{Mean and standard deviation fitness values of $F_4$, \textbf{RSME}.}
	\label{tab:ResultsObj3}

\vspace{2em}
\begin{tabular}{|l|c|c|c|c|}
\hline
Patient & Grammar \gdiez & Grammar \gonce & Grammar \gdoce & Grammar \gtrece \\ 
\hline 
\joywilson & $0.1128_{0.0108} $ & $\mathbf{0.0691_{0.0064}} $ & $0.1176_{0.0124} $ & $0.0706_0.0055$ \\ 
\howard & $\mathbf{0.2176_{0.0040}} $ & $0.2410_{0.0567} $ & $0.2186_{0.0991} $ & $0.2190_{0.0528}$ \\  
\steven & $0.1335_{0.0186} $ & $0.1153_{0.0268} $ & $0.1380_{0.0208} $ & $\mathbf{0.1119_{0.0230}}$ \\
\elizabeth & $0.2196_{0.0072} $ & $0.2398_{0.0112} $ & $\mathbf{0.2152_{0.0162}} $ & $0.2375_{0.0147}$ \\
\lizzy & $\mathbf{0.0908_{0.0109}} $ & $0.0952_{0.0084} $ & $0.0942_{0.0088} $ & $0.0953_{0.0091}$ \\
\hline 
\end{tabular}
	\caption{Mean and standard deviation fitness values of $F_5$, \textbf{MAD}.}
	\label{tab:ResultsObj4}
\end{footnotesize}
	
\end{center}
\end{table}

The mean and standard deviation fitness values for the average error objective are shown in Table \ref{tab:ResultsObj1}. Grammar \gtrece obtains here two out of five best results, being close to the best one in patient \steven. Grammar \gonce also performs quite well, obtaining second best results where Grammar \gtrece wins.


Table \ref{tab:ResultsObj2} displays the mean and standard deviation fitness for the maximum error objective. Here, Grammars \gdoce and \gtrece obtain two best results each. However Grammar \gtrece is better for patient \steven, where Grammar \gonce wins.


The mean and standard deviation fitness values for RSME objective are shown in Table \ref{tab:ResultsObj3}. We can see here that Grammar \gonce obtains four out of five best results. Grammar \gtrece obtains four second best values, which is also a good performance.


Table \ref{tab:ResultsObj4} presents mean and standard deviation fitness values for objective MAD. Here we found that Grammar \gdiez\ obtains two best results and one second best.


In general, the best objective-grammar combination will be the one that obtains best average fitness for any input data. In our experiments, none of the combinations reached this goal. Grammar \gonce is close to it in least squares and RSME objectives, but the other grammars perform well in the other objectives.

Therefore, in order to complete this analysis, we next compare the quality of the solutions obtained for each one of the patients. Breaking down these results by objective will give the idea of which fitness function could be better.

\subsection{Analysis and Test Phase}

Once we have seen the overall performance of the grammars and fitness functions, we analyze the results for our input datasets.

For each patient, we have calculated the percentage that the average error of each simulation run represents in the range of the patient glucose values. Hence, we have obtained the mean and standard deviation of the percentage average error for the 30 runs of each grammar and objective combination.

Then, we have run the test phase for the best grammar and objective combination on each patient training. To this aim we needed different inputs which, in this case, consisted on different 24-hours datasets generated with AIDA simulator. Hence, in order to obtain variations on the actual glucose values, we changed the parameters in the simulator, varying carbohydrates and/or insulin units trying to represent realistic situations like bigger or smaller meals and little changes in the insulin doses.

Next, we analyze the results for each patient dataset.

\subsubsection{\joywilson}

Table \ref{tab:statsjoy} shows the mean and standard deviation of percentage average error for \joywilson. As seen, the best combination of grammar and objective is Grammar \gonce optimizing MAD. Notice that the best average error is not obtained optimizing the average error objective, which also happens for Grammar \gdiez and least squares objective. 

\begin{table}\footnotesize
\begin{center}
\begin{tabular}{|l|c|c|c|c|}
\hline
Objective & Grammar \gdiez & Grammar \gonce & Grammar \gdoce & Grammar \gtrece \\
\hline
Least Squares &$16.18_{3.7}$ & $11.37_{1.38}$ & $18.04_{2.23}$ & $11.6_{1.26}$\\
Average Error & $16.68_{2.83}$ & $10.9_{1.18}$ & $16.67_{2.56}$ & $10.79_{1.12}$\\
Max. Error &$19.22_{2.56}$ & $15.29_{1.05}$ & $19.37_{1.96}$ & $15.11_{0.76}$\\
RSME & $17.21_{2.84}$ & $11.24_{0.87}$ & $18.04_{2.05}$ & $11.61_{1.36}$\\
MAD &$17.72_{1.57}$ & $\mathbf{10.68_{0.99}}$ & $18.24_{2.03}$ & $10.91_{0.82}$\\
\hline 
\end{tabular}
\end{center}
	\caption{Mean and standard deviation of percentage average error, patient \joywilson.}
	\label{tab:statsjoy}
\end{table}

\begin{figure}[htbp]
   \centering
	\begin{subfigure}{0.5\textwidth}   
		\centering
	    \includegraphics[width=\textwidth]{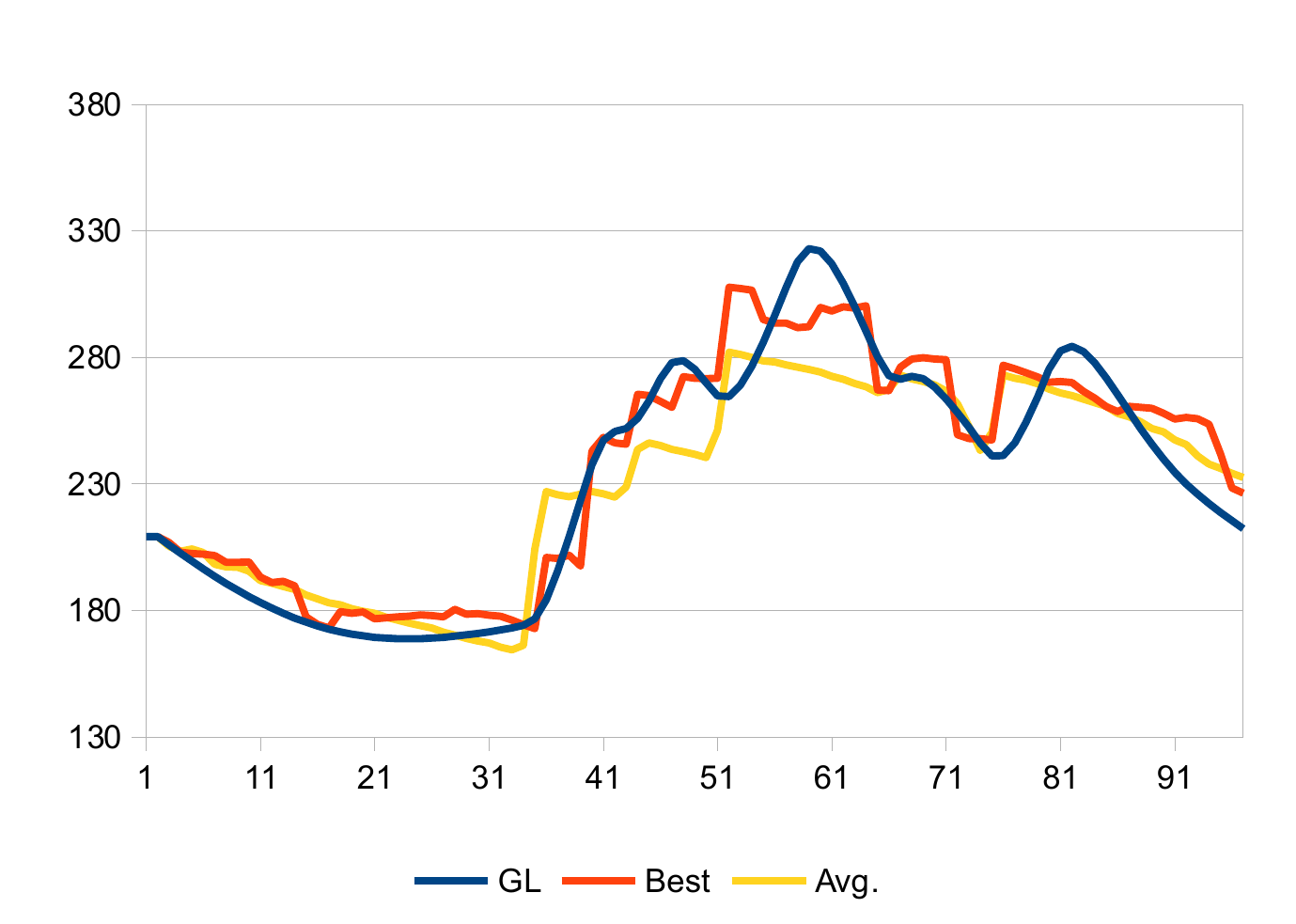} 
	    \caption{Training.}
    	\label{fig:graphjoya}
    \end{subfigure}%
	\begin{subfigure}{0.5\textwidth}   
		\centering
	     \includegraphics[width=\textwidth]{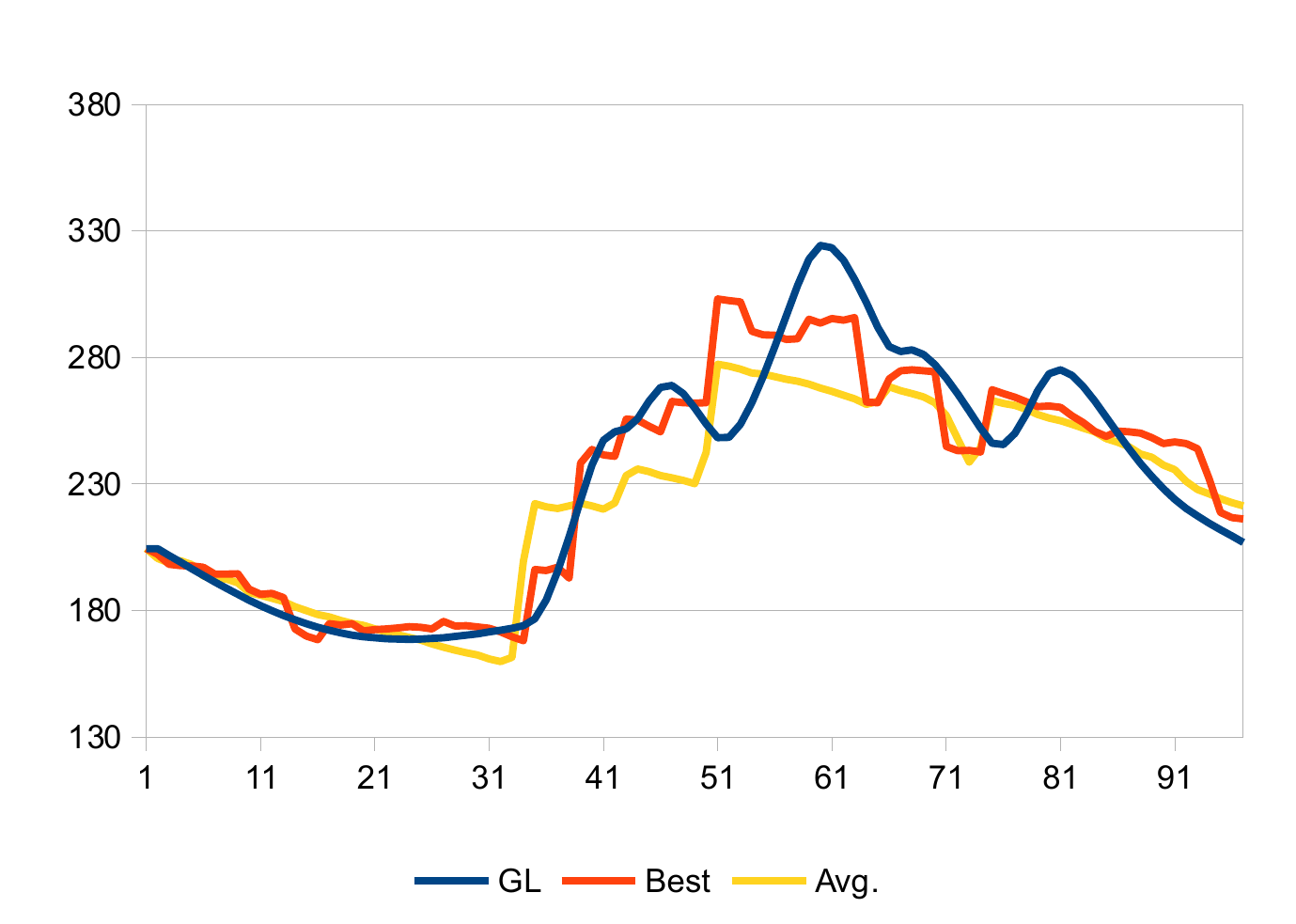} 
	    \caption{Test.}
    	\label{fig:graphjoyb}
    \end{subfigure}%
   \caption{Best combination for \joywilson: Grammar \gonce and MAD.}
   \label{fig:graphjoy}
\end{figure}

Figure \ref{fig:graphjoya} shows the glucose values obtained with the best grammar-objective combination of the training phase for \joywilson. The actual glucose curve of the patient (in blue), the glucose value generated with the best solution of this combination (in red) and the glucose value generated with the average of the 30 solutions (in yellow) are displayed in the figure. The best solution obtained a percentage average error value of 7.37\%, and its expression was the following:

\begin{scriptsize}
\[GL(k+1)=GL(k)+CH(k-1)-cos(IL(k-1))+tan(exp(IL(k-1)+cos(tan(exp(exp(cos(k)))))))\]
\end{scriptsize}

The test phase for \joywilson\ was run with a dataset obtained by decreasing the 10 AM snack from 20 to 15 carbohydrate units, increasing lunch from 40 to 45 carbohydrate units and decreasing dinner from 30 to 25 carbohydrate units in the simulator. Insulin values were not modified. Figure \ref{fig:graphjoyb} shows the actual glucose value given by the simulator, as well as the values given by the best and average solutions obtained from training. As seen, the first third of the best solution is close to the actual glucose, while in the rest the gap is bigger than in training. The best solution obtained a percentage average error value of 7.41\% in the test phase.

\subsubsection{\howard}

Statistics for \howard\ are shown in Table \ref{tab:statshoward}. Here, the best percentage average error is obtained with Grammar \gtrece optimizing the average error. This objective obtains the best value also for Grammar \gdiez, while for Grammar \gonce is not as good as least squares and RSME.

\begin{table}\footnotesize
\begin{center}
\begin{tabular}{|l|c|c|c|c|}
\hline
Objective & Grammar \gdiez & Grammar \gonce & Grammar \gdoce & Grammar \gtrece \\
\hline
Least Squares & $14.25_{0.58}$ & $14.07_{4.84}$ & $14.68_{2.21}$ & $14.9_{4.96}$\\
Average Error & $13.93_{0.65}$ & $15.45_{3.72}$ & $13.97_{1.01}$ & $\mathbf{13.34_{4.07}}$\\
Max. Error & $21.21_{2.34}$ & $19.27_{4.52}$ & $19.25_{3.26}$ & $21.18_{3.35}$\\
RSME & $14.06_{0.56}$ & $14.9_{4.42}$ & $14.15_{1.69}$ & $16.36_{4.44}$\\
MAD & $14.18_{0.28}$ & $17.09_{4.99}$ & $13.83_{2.75}$ & $15.26_{4.57}$\\
\hline 
\end{tabular}
\end{center}
	\caption{Mean and standard deviation of percentage average error, patient \howard.}
	\label{tab:statshoward}
\end{table}

\begin{figure}[htbp]
   \centering
	\begin{subfigure}{0.5\textwidth}   
		\centering
	    \includegraphics[width=\textwidth]{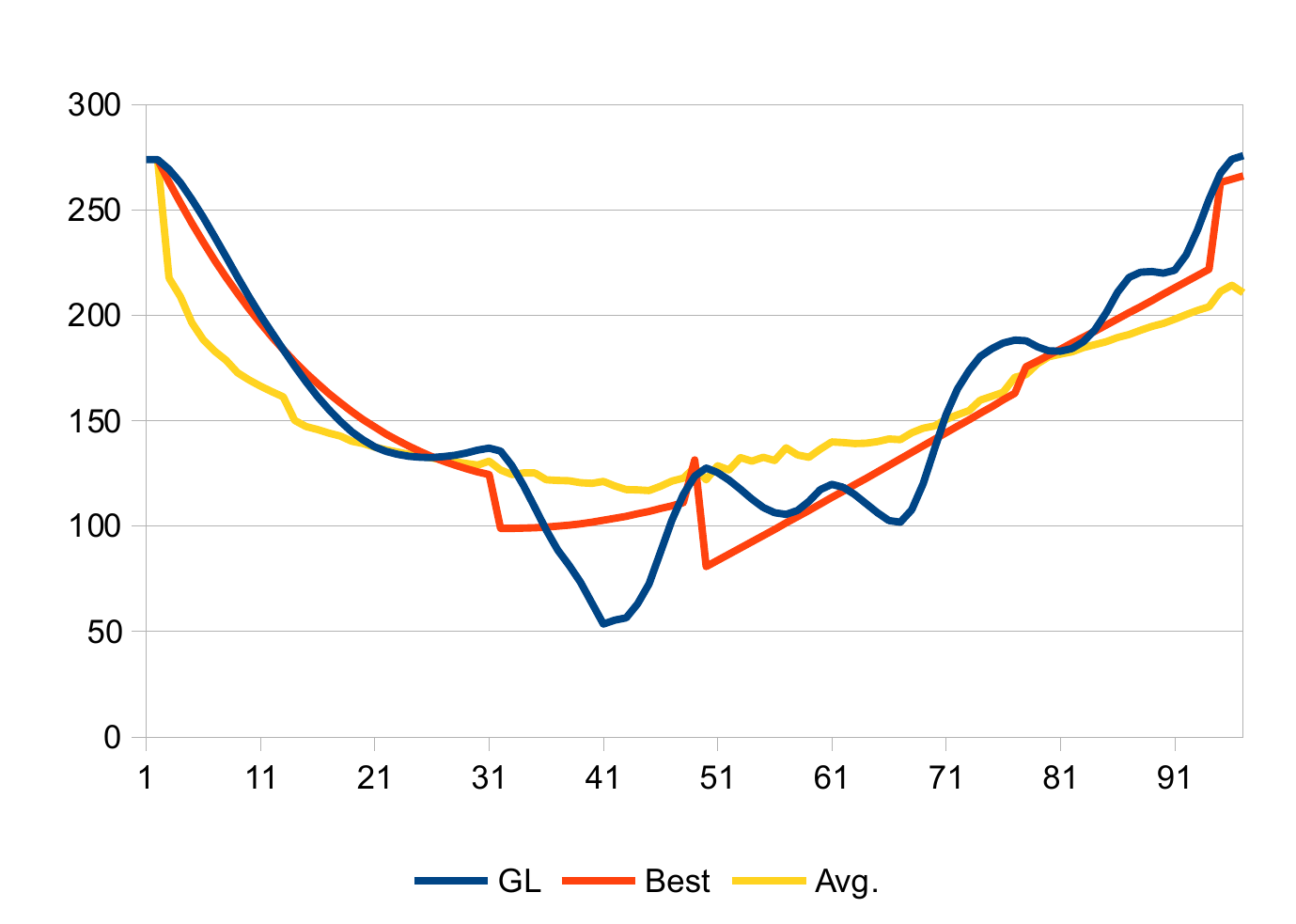} 
	    \caption{Training.}
    	\label{fig:graphhowarda}
    \end{subfigure}%
	\begin{subfigure}{0.5\textwidth}   
		\centering
	     \includegraphics[width=\textwidth]{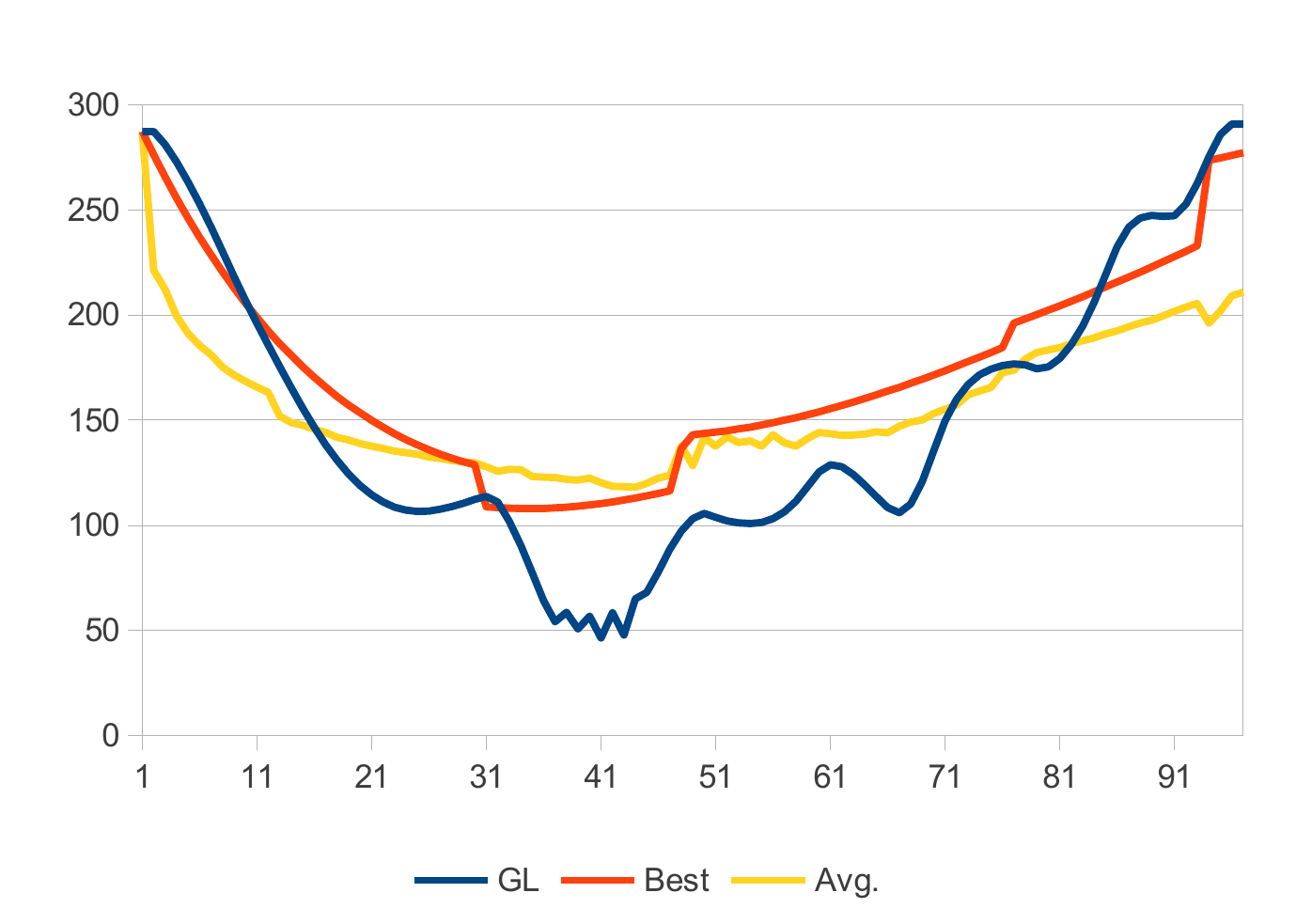} 
	    \caption{Test.}
    	\label{fig:graphhowardb}
    \end{subfigure}%
   \caption{Best combination for \howard: Grammar \gtrece and average error.}
   \label{fig:graphhoward}
\end{figure}

Figure \ref{fig:graphhowarda} shows the actual glucose value obtained from the simulator as well as the best solution and the average of the runs for Grammar \gtrece optimizing average error. It can be seen that the average glucose does not follow the actual glucose as well as the best solution. This is because the variability of the solutions in this combination, expressed in the value of the standard deviation, 4.07\%, which is a little high. The best solution obtained a percentage average error value of 6.62\%, and its expression was the following:

\begin{scriptsize}
\[GL(k+1)=\frac{41.57*GL(k)}{43.24}+k*cos(exp(exp(sin(IS(k-1))-IL(k)-cos(IS(k-1)*IS(k-1)))))\]
\end{scriptsize}

The test phase for \howard\ consisted on varying the insulin doses on one unit less or one unit more. The change on the actual glucose that is more relevant in the test phase is the decrease around the measure number 20, as shown in Figure \ref{fig:graphhowardb}. This was caused because we increased the value of the long effect insulin from 6 to 7 units. The other changes on the insulin were not so evident on the actual glucose. However, given that the best model is very dependent on the insulin, those changes separated the best solution curve in relation to the training. As a result, the best solution obtained a percentage average error value of 11.33\% in the test phase.

%
%

\subsubsection{\steven}

The percentage average error values for \steven\ are shown in Table \ref{tab:statssteven}. The best result is obtained with Grammar \gonce optimizing the average error. Besides, we can see that Grammar \gonce obtains the best average in all objectives but in the case of maximum error, where it also is very close indeed. 

\begin{table}\footnotesize
\begin{center}
\begin{tabular}{|l|c|c|c|c|}
\hline
Objective & Grammar \gdiez & Grammar \gonce & Grammar \gdoce & Grammar \gtrece \\
\hline
Least Squares & $18.91_{3.15}$ & $14.71_{1.73}$ & $20.58_{3.59}$ & $15.87_{2.33}$\\
Average Error & $19.26_{2.55}$ & $\mathbf{14.01_{1.91}}$ & $19.25_{3.29}$ & $14.86_{2.6}$\\
Max. Error & $22.28_{0.98}$ & $18.42_{2.35}$ & $22.34_{1.82}$ & $18.23_{1.97}$\\
RSME & $19.17_{2.94}$ & $14.61_{2.14}$ & $19.66_{3.24}$ & $15.38_{2.21}$\\
MAD & $18.56_{2.55}$ & $15.66_{3.52}$ & $18.95_{2.93}$ & $14.96_{3.25}$\\
\hline 
\end{tabular}
\end{center}
	\caption{Mean and standard deviation of percentage average error, patient \steven.}
	\label{tab:statssteven}
\end{table}

\begin{figure}[htbp]
   \centering
	\begin{subfigure}{0.5\textwidth}   
		\centering
	    \includegraphics[width=\textwidth]{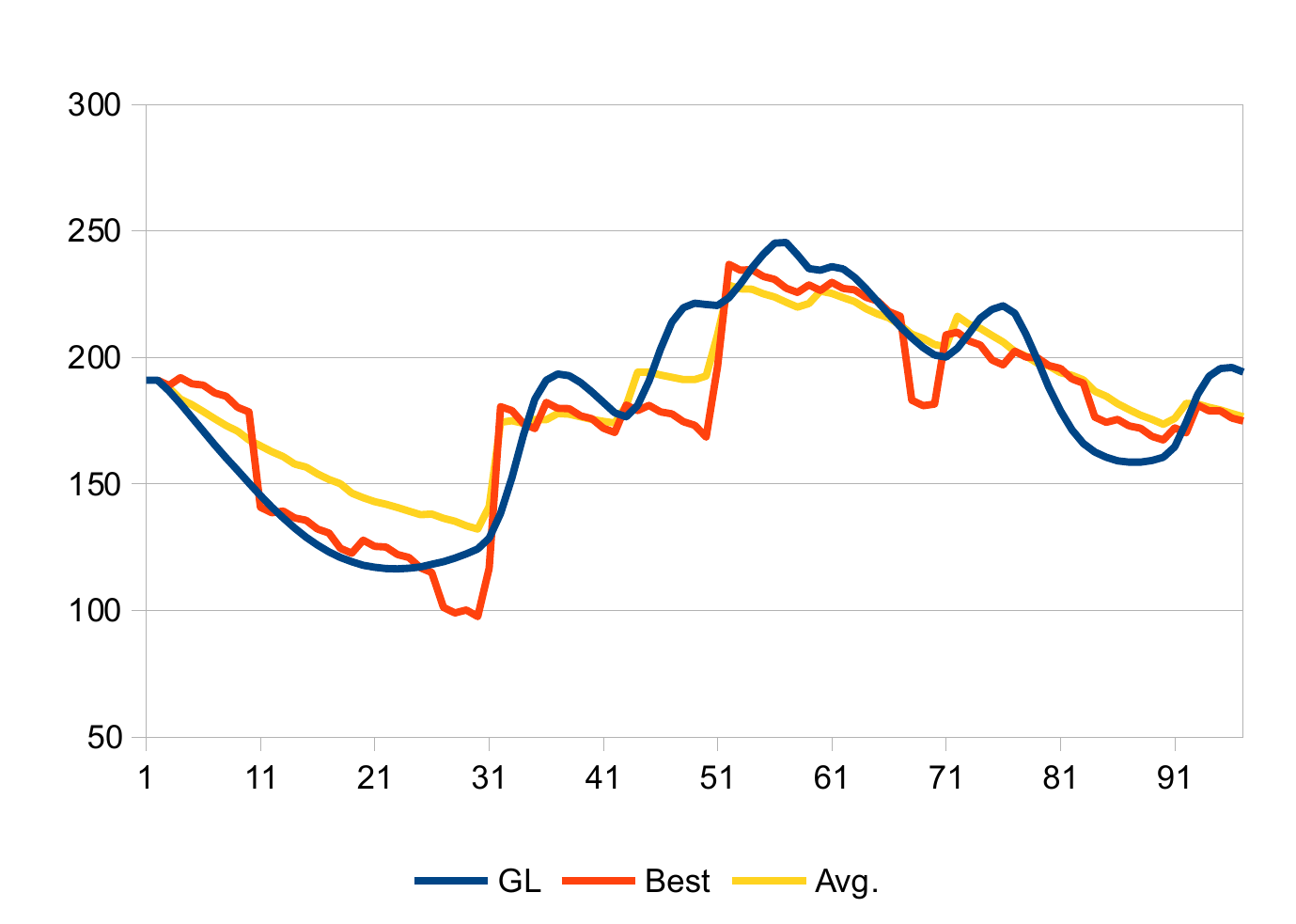} 
	    \caption{Training.}
    	\label{fig:graphstevena}
    \end{subfigure}%
	\begin{subfigure}{0.5\textwidth}   
		\centering
	     \includegraphics[width=\textwidth]{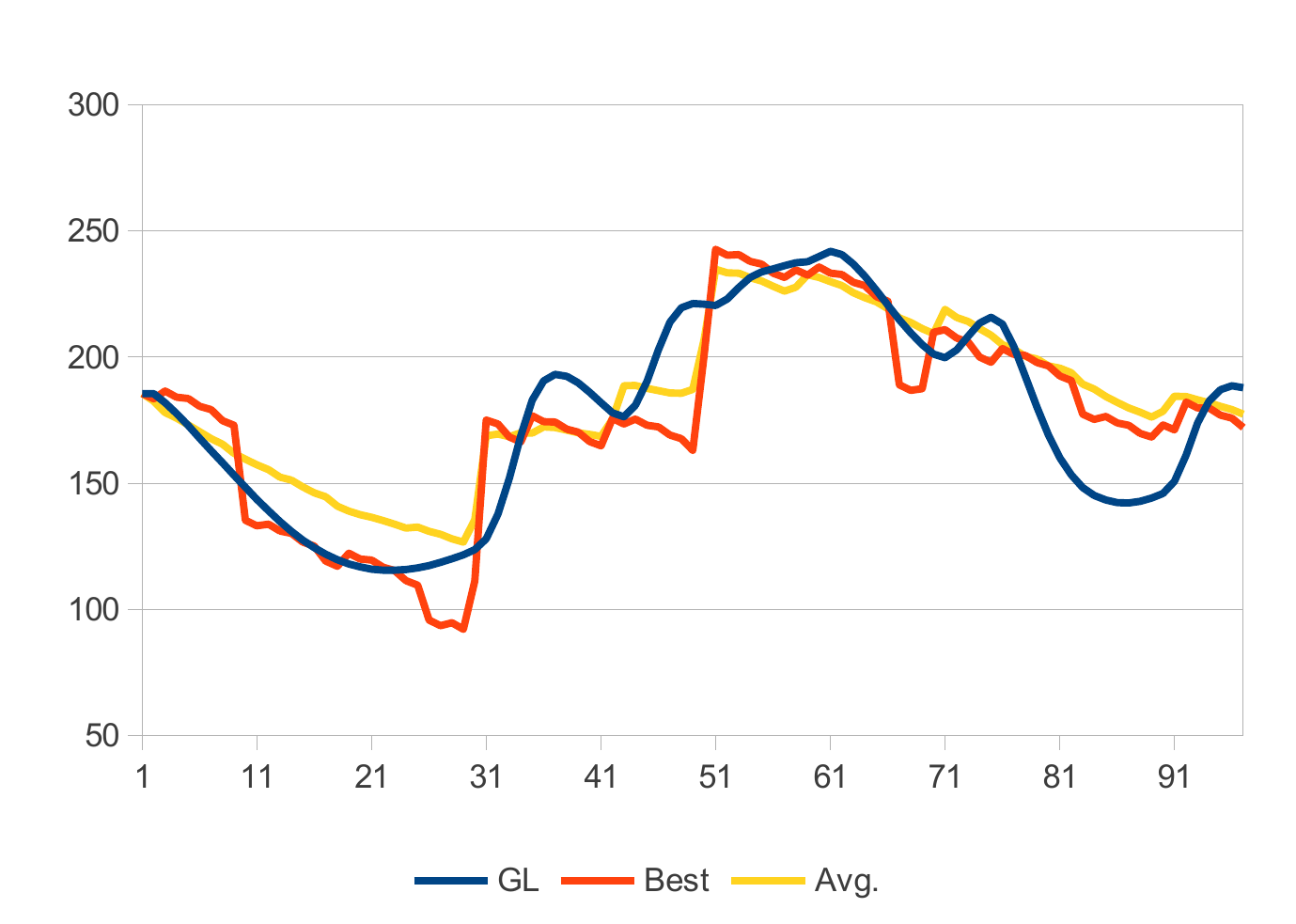} 
	    \caption{Test.}
    	\label{fig:graphstevenb}
    \end{subfigure}%
   \caption{Best combination for \steven: Grammar \gonce and average error.}
   \label{fig:graphsteven}
\end{figure}

The plots for the glucose in the training phase are displayed in Figure \ref{fig:graphstevena}. Both best and average solutions have a shape similar to the actual glucose. However, despite the shapes are similar, due to a short range of glucose values, the best solution obtained a percentage average error value of 9.83\%. Notice that this percentage is calculated with respect to the range of glucose of each patient. The expression of the best solution was the following:

\begin{footnotesize}
\[GL(k+1)=GL(k)+CH(k)-tan(tan(sin(exp(tan(IS(k-1))))))-tan(k*14.33)\]
\end{footnotesize}

In the test phase of \steven\ we increased the lunch carbohydrates from 30 to 40 units and the short insulin from 3 to 4 units. As seen in Figure \ref{fig:graphstevenb}, this caused a lower peak between measures 50 and 60 in the actual glucose. We also decreased the 5 PM dinner from 30 to 25 carbohydrate units maintaining the insulin doses. Hence, we see that the end of the glucose curve is lower in the test plot. In this patient, these variations are not captured by the best and average models. In fact, the best solution obtained a percentage average error value of 11.53\% in the test phase. 

%
%

\subsubsection{\elizabeth}

Table \ref{tab:statselizabeth} shows the percentage average error values for \elizabeth. The best combination is Grammar \gdoce with RSME objective. Once again, minimizing the average error objective does not obtain the best average results. Besides, in this patient grammars \gdiez and \gdoce obtain better results than the others. These grammars may consider any previous carbohydrate or insulin values, while grammars \gonce and \gtrece may consider just the two previous data. This behavior is caused by the shape of the actual glucose values, that are similar to sawtooth in the middle, being very difficult to imitate. 

\begin{table}\footnotesize
\begin{center}
\begin{tabular}{|l|c|c|c|c|}
\hline
Objective & Grammar \gdiez & Grammar \gonce & Grammar \gdoce & Grammar \gtrece \\
\hline
Least Squares & $17.24_{1.46}$ & $18.07_{1.34}$ & $16.86_{1.35}$ & $18.23_{1.38}$\\
Average Error & $16.91_{1.28}$ & $17.79_{1.54}$ & $16.6_{1.07}$ & $17.34_{1.67}$\\
Max. Error & $20.46_{1.37}$ & $20.57_{0.92}$ & $20.24_{1.33}$ & $20.59_{0.85}$\\
RSME &$17.36_{1.21}$ & $17.79_{1.6}$ & $\mathbf{16.46_{1.6}}$ & $17.34_{1.78}$\\
MAD & $17.47_{0.94}$ & $19.2_{1.14}$ & $16.97_{1.38}$ & $18.96_{1.56}$\\
\hline 
\end{tabular}
\end{center}
	\caption{Mean and standard deviation of percentage average error, patient \elizabeth.}
	\label{tab:statselizabeth}
\end{table}

\begin{figure}[htbp]
   \centering
	\begin{subfigure}{0.5\textwidth}   
		\centering
	    \includegraphics[width=\textwidth]{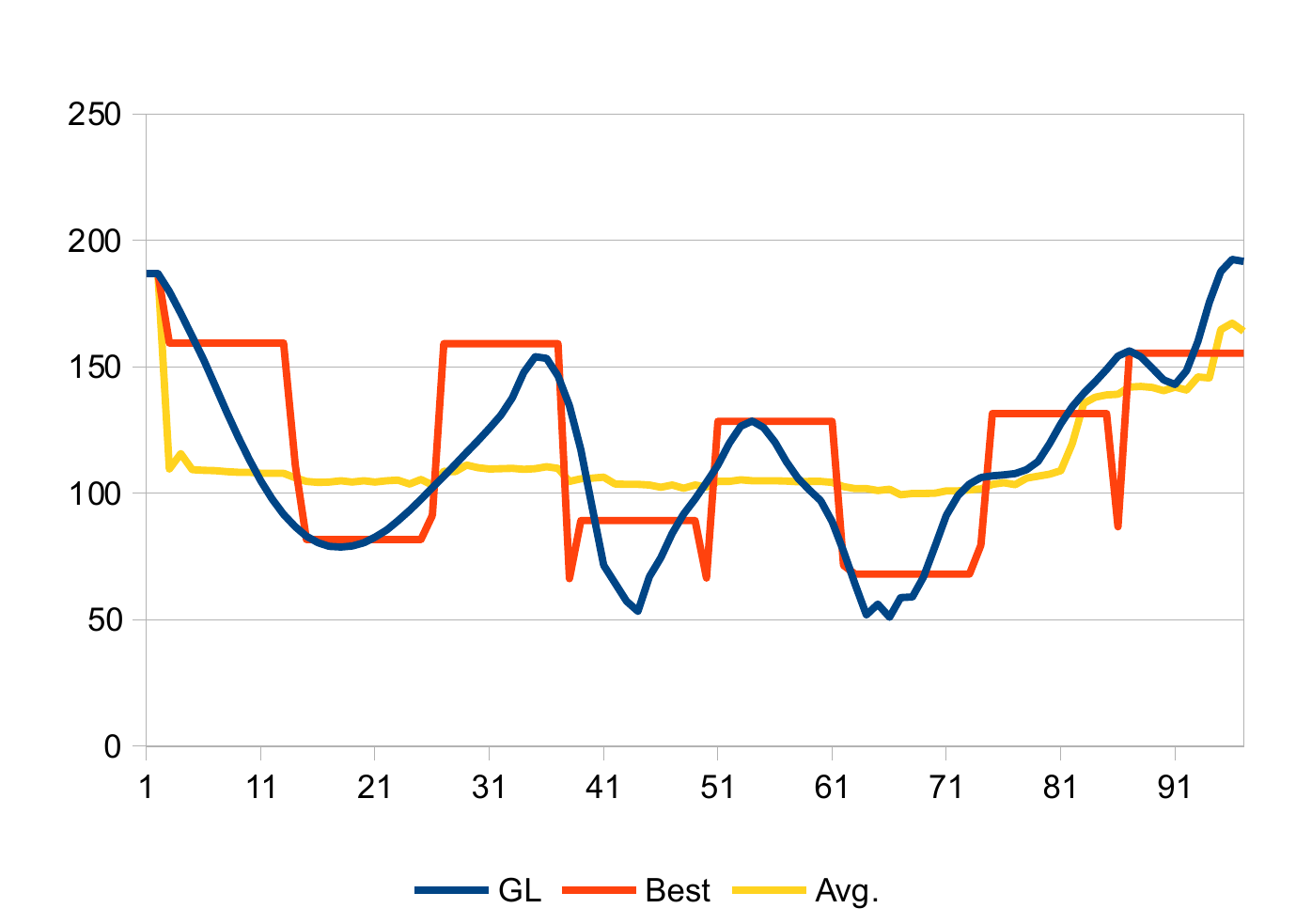} 
	    \caption{Training.}
    	\label{fig:graphelizabetha}
    \end{subfigure}%
	\begin{subfigure}{0.5\textwidth}   
		\centering
	     \includegraphics[width=\textwidth]{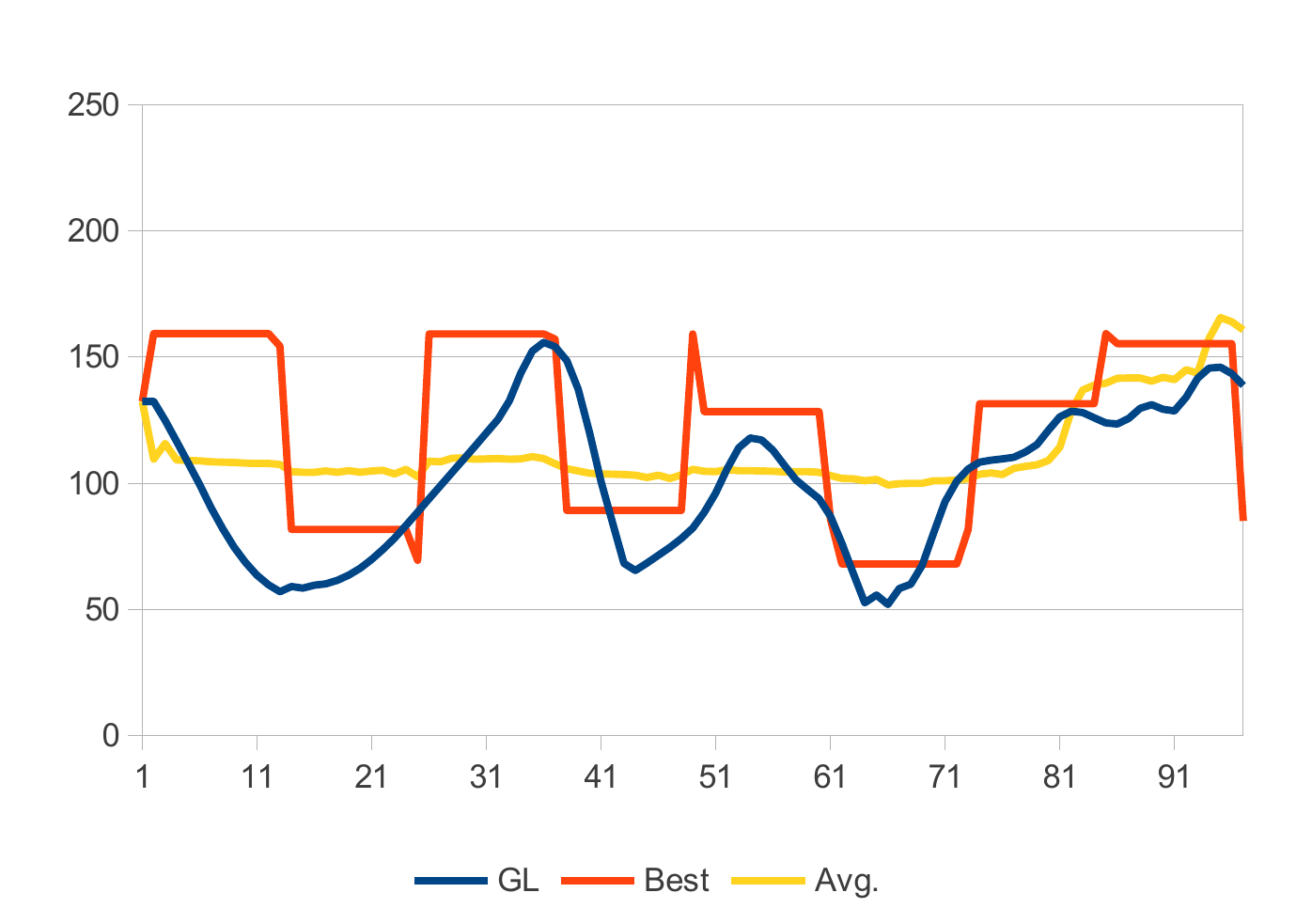} 
	    \caption{Test.}
    	\label{fig:graphelizabethb}
    \end{subfigure}%
   \caption{Best combination for \elizabeth: Grammar \gdoce and RSME.}
   \label{fig:graphelizabeth}
\end{figure}

Figure \ref{fig:graphelizabetha} shows the special shape of the actual glucose, as well as the more special squared shape of the best solution given, and the plain average solution. The latter is similar to an interpolation, and does not help so much. The best solution obtained a percentage average error value of 13.46\%. The expression of the best solution was the following:

\begin{footnotesize}
\[GL(k+1)=cos(GL(k-11))*46.98+15.45+96.93\]
\end{footnotesize}

The expression in this case is very constant, and the cosine depends on the glucose estimated for a time step three hours ago. For the test phase we decreased from 6 to 5 units the short effect insulin at breakfast, decrease the 10 AM snack from 20 to 10 carbohydrate units, increase dinner from 40 to 50 carbohydrate units, increase short effect insulin at dinner from 3 to 4 units, and decrease the 10 PM snack from 20 to 10 carbohydrate units. These changes modified both the actual glucose in test and the best solution values, as seen in Figure \ref{fig:graphelizabethb}. However, this is a difficult dataset where the best solution obtained a percentage average error value of 26.40\% in the test phase.  

%

\subsubsection{\lizzy}

Looking at one objective of the results for patient \lizzy, presented in Table \ref{tab:statslizzy}, we see that there are not very significative differences between grammars. However, the best result is obtained with Grammar \gdiez optimizing the average error objective.

\begin{table}\footnotesize
\begin{center}
\begin{tabular}{|l|c|c|c|c|}
\hline
Objective & Grammar \gdiez & Grammar \gonce & Grammar \gdoce & Grammar \gtrece \\
\hline
Least Squares &$16.62_{1.93}$ & $16.98_{1.44}$ & $17.19_{1.72}$ & $17.17_{2.46}$\\
Average Error & $\mathbf{16.11_{1.98}}$ & $16.84_{1.73}$ & $16.46_{1.88}$ & $16.99_{1.72}$\\
Max. Error & $25.24_{2.66}$ & $22.56_{2.6}$ & $24.15_{3.47}$ & $22.36_{4.37}$\\
RSME & $17.17_{1.43}$ & $16.71_{1.63}$ & $17.12_{1.92}$ & $16.8_{2.24}$\\
MAD & $16.32_{1.95}$ & $17.19_{1.47}$ & $16.87_{1.75}$ & $17.16_{1.74}$\\
\hline 
\end{tabular}
\end{center}
	\caption{Mean and standard deviation of percentage average error, patient \lizzy.}
	\label{tab:statslizzy}
\end{table}

\begin{figure}[htbp]
   \centering
	\begin{subfigure}{0.5\textwidth}   
		\centering
	    \includegraphics[width=\textwidth]{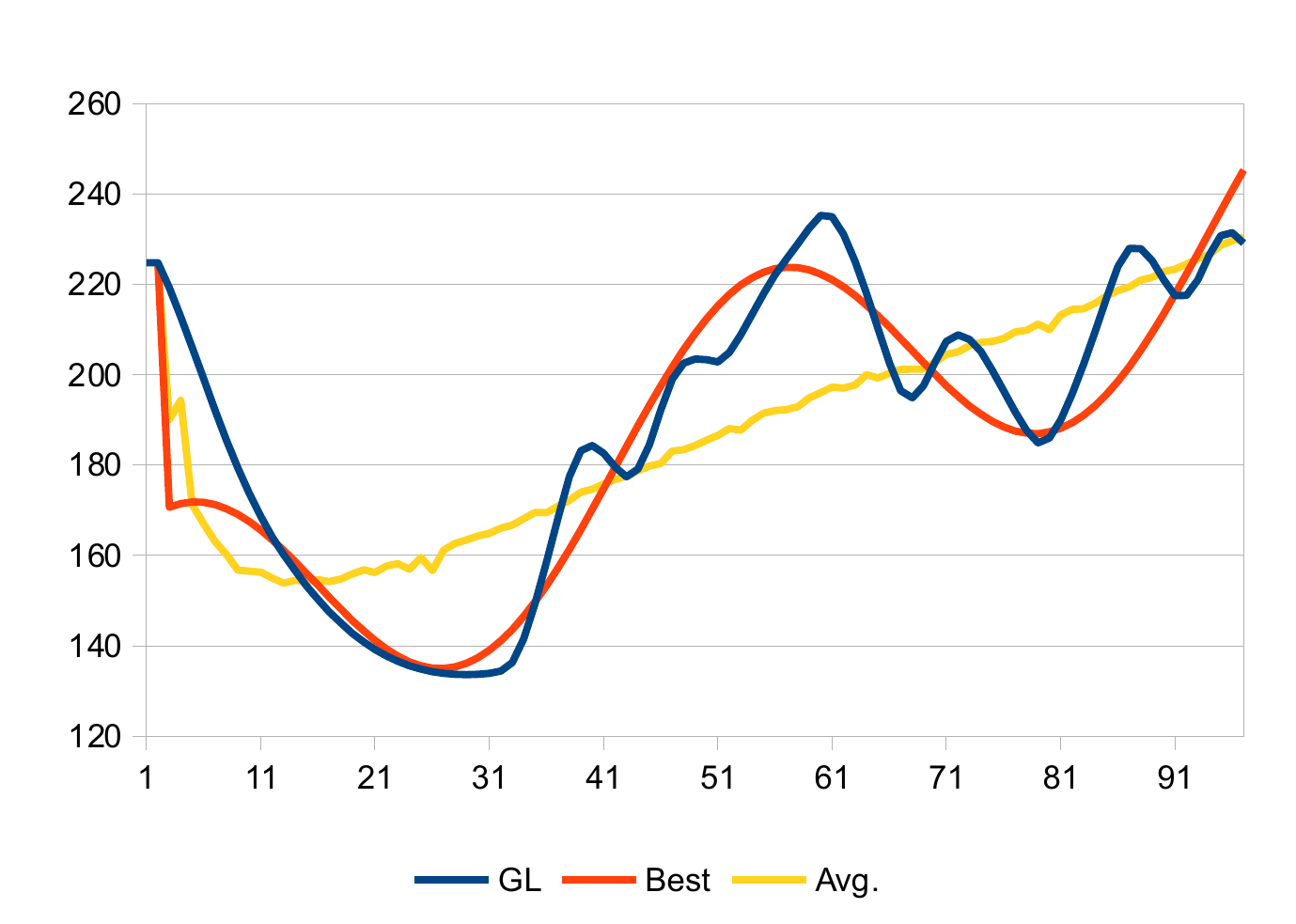} 
	    \caption{Training.}
    	\label{fig:graphlizzya}
    \end{subfigure}%
	\begin{subfigure}{0.5\textwidth}   
		\centering
	     \includegraphics[width=\textwidth]{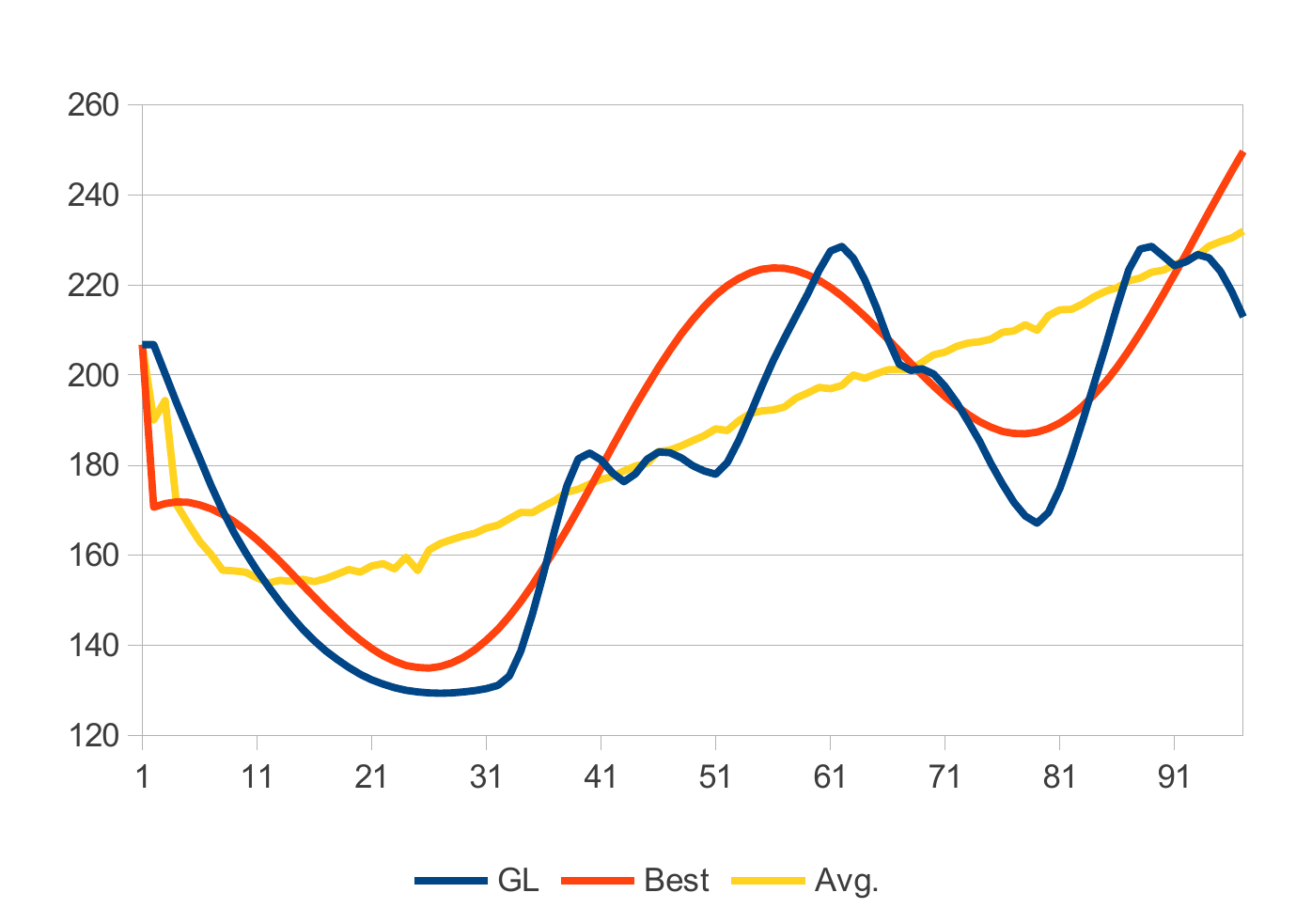} 
	    \caption{Test.}
    	\label{fig:graphlizzyhb}
    \end{subfigure}%
   \caption{Best combination for \lizzy: Grammar \gdiez and average error.}
   \label{fig:graphlizzy}
\end{figure}

Actual glucose values for training and best and average solutions are displayed in Figure \ref{fig:graphlizzya}.  As seen, the best solution presents a similar shape as the actual glucose, while the average, once again, behaves like an interpolation. The best solution obtained a percentage average error value of 9.34\%. The expression of the best solution was the following:

\begin{footnotesize}
\[GL(k+1)=(89.91+k)+50.51-\frac{cos(37.82*k)}{tan(84.79)}\]
\end{footnotesize}

In this case it is clear that the best solution will obtain worse results in the test phase because its expression only depends on $k$. So, for the test we reduced the snacks in from 20 to 10 carbohydrate units, and increased lunch and dinner in 5 carbohydrate units. As seen in Figure \ref{fig:graphlizzyhb} the best solution remains the same than in training, while the actual glucose value changes. The best solution obtained a percentage average error value of 11.8\% in this test phase.  

%
%

\subsection{Discussion}

The results obtained in our experiments raised several issues related both with grammars and with objective functions.

Regarding the grammars, we have found that grammars \gonce and \gtrece, which consider the data previous to each time step, behave quite well in all the objectives under study. Therefore, the intuition that recent values recall the previous history (glucose, meals, insulin) is validated here. In addition, the expressions obtained as best solutions with grammars \gonce and \gtrece depend on carbohydrate and insulin units. Therefore, these expressions consider the input values that a patient can collect and, as a consequence, the expressions may behave well in test phases.

On the other hand, despite of the average error good values, grammars \gdiez and \gdoce have provided useless expressions that are less parameterized with the inputs of the patient. Moreover, the average solutions in these cases tend to interpolate the glucose values, while in the other grammars the average is similar to the actual value.

In terms of objectives, and given that we use the average error as a quality measure, this could be the best objective to choice. Nevertheless, the most of the studied objectives behave quite well, with the exception of the maximum error one. In fact, minimizing this objective does not tend to minimize the average error so much, which can be viewed as an opposite objective useful for future multi-objective optimizations.

%% file: conclusions.tex
In this paper we propose an evolutionary method based on GE that automatically obtains custom models for blood glucose levels in diabetic patients. Up to our knowledge, this is the first proposal where GE is applied to obtain glucose models in diabetics.

The main advantages of our method are: (1) the model is obtained as a custom expression for each patient, which improves the individual treatment of a diabetic person; (2) the training dataset can be easily collected by a patient or by a simple system because models require values of previous glucose measures, carbohydrate units ingested and insulin doses injected; (3) this method may be integrated in a progressive optimization system where the model is generated and stored and, after several days of data gathering, the model can be updated using the new dataset.

In our work, we have studied four different grammars and five different objective functions for our optimization scheme on five in-silico patients. The grammars incorporated some knowledge about the problem, trying to limit the search space of the algorithm. We have concluded that grammars which consider previous data that are close to the current time step are better than those able to select any previous data. That is, the most recent data are more valuable than the past ones. In addition, these grammars obtained more useful models because their expressions depend on almost all the involved variables. Besides, we saw that optimizing the average error objective obtains the best results, as well as we identified that the maximum error is an opposite objective that could be considered in future multi-objective optimizations.

Once the training phase finished, we selected the best model expressions for each patient and run the test phase with a different dataset. The results showed a mean percentage average error of 13.69\% for the best models in the test phase. In addition, the best models predicted quite well the dangerous situations of hyper and hypoglucemias for all the patients.

In the future, we expect to manage datasets from real patients, which will allow the study of new variables in the models like stress or exercise. This will require the refinement of the grammars. In addition, we will consider the multiobjective optimization with both average and maximum error objectives. We will also consider to integrate fuzzy regression into the GP process \cite{Chan2010} \cite{Chan2011}.